\documentclass{CUP-JNL-NLP}
\usepackage{appendix}

\usepackage{amsmath}
\usepackage{graphicx}
\usepackage{pdflscape}
\usepackage{xcolor}
\usepackage{natbib}
\ifpdf
\usepackage{epstopdf}
\fi
\usepackage{multirow}
\usepackage{amsthm} 
\usepackage{enumitem} 
\usepackage{caption}

\newenvironment{exm}[1][]{%
    \refstepcounter{example}%
    \par\medskip%
    \noindent\textbf{Example \theexample. #1}%
    \par\medskip%
}{%
    \par\medskip%
}

\def\aulastand{\ and\nobreakspace{}}
\def\auandname{and\nobreakspace{}}

\begin{document}
\label{firstpage}

\lefttitle{Assessing and Refining ChatGPT's Performance}
\righttitle{Natural Language Processing}

\papertitle{Article}

\jnlPage{\pageref{firstpage}}{\pageref{lastpage}}
\jnlDoiYr{2026}
\doival{10.1017/xxxxx}

\title{Assessing and Refining ChatGPT's Performance in Identifying Targeting and Inappropriate Language: A Comparative Study}

\begin{authgrp}
    \author{Baran Barbarestani}
        \affiliation{Vrije Universiteit Amsterdam,\\
        Computational Linguistics and Text Mining Lab\\
        \email{b.barbarestani@vu.nl}}
    \author{Isa Maks}
        \affiliation{Vrije Universiteit Amsterdam,\\
        Computational Linguistics and Text Mining Lab\\
        \email{isa.maks@vu.nl}}
    \author{Piek Vossen}
        \affiliation{Vrije Universiteit Amsterdam,\\
        Computational Linguistics and Text Mining Lab\\
        \email{p.t.j.m.vossen@vu.nl}}
\end{authgrp}
\history{(Received xx xxx xxx; revised xx xxx xxx; accepted xx xxx xxx)}

\begin{abstract}
Attention: This paper includes instances of hateful content for research purposes.

This study evaluates the effectiveness of ChatGPT, an advanced AI model for natural language processing, in identifying targeting and inappropriate language in online comments. With the increasing challenge of moderating vast volumes of user-generated content on social network sites, the role of AI in content moderation has gained prominence. We compared ChatGPT's performance against crowd-sourced annotations and expert evaluations to assess its accuracy, scope of detection, and consistency. Our findings show that ChatGPT performs well in detecting inappropriate content, and that almost the entire gain from iterative prompt refinement came from a single change --- simplifying the label scheme and moving to GPT-4 --- after which further adjustments to prompt structure and sampling temperature had little effect. Its performance on targeting language was weaker and less stable, with higher false positive rates than expert judgments before refinement and higher false negative rates after a change of model. This study contributes to the field by demonstrating the potential of AI models like ChatGPT to enhance automated content moderation systems while also identifying areas for further improvement. The results underscore the importance of careful prompt design and model selection, and of contextual understanding, in supporting automated moderation and mitigating harmful online behavior.
\end{abstract}

\maketitle

\section{Introduction}
\subsection{Motivation and Context}
\par Social network sites (SNSs), like Facebook, Twitter, and YouTube, facilitate online interaction and social relationships, bridging geographical distances and connecting individuals, both known and unknown, through mediated communication \citep{thompson2020mediated}. However, the prevalence of Internet trolling, characterized by antisocial behavior, poses significant challenges for these platforms \citep{paakki2021disruptive}. While educational efforts and dialogues between adults and youth are underway to address this issue, technical solutions within social network software are increasingly recognized as vital for identifying and mitigating harmful conduct \citep{dinakar2012common}.
\par Artificial intelligence (AI) tools can aid human moderators, improving content filtering and efficiency \citep{haque2024exploring}. However, scalability remains a challenge in content moderation, as the vast volume of online content exceeds human moderation capacity \citep{schluger2022proactive}. While AI models show promise across various tasks due to extensive data availability \citep{zhang2022artificial}, they face technical hurdles such as data set shifts, overfitting, bias, and generalization issues \citep{yin2021role}, limiting their practical application. ChatGPT, an AI model for conversation created by OpenAI and first released in November 2022, represents a significant advancement in natural language processing (NLP) \citep{goar2023conversational}. Its ability to understand and generate human-like text positions it as a candidate for enhancing automated content moderation systems. This study aims to evaluate ChatGPT's performance in identifying and classifying targeting and inappropriate language, comparing its effectiveness with crowd-sourced annotations and expert evaluations.
\subsection{Objectives}
\par The primary objectives of this study are twofold:
\begin{enumerate}
    \item To assess ChatGPT's performance: We aim to evaluate how well ChatGPT identifies targeting and inappropriate language in online comments compared to crowd-sourced and expert annotations. This involves analyzing its scope of detection, accuracy, and consistency in identifying problematic content.
    \item To improve model accuracy: By iteratively refining the model's prompts and configurations, we seek to enhance ChatGPT's alignment with human judgments. This involves testing different versions of the model and analyzing how changes in prompt design and model parameters affect performance.
\end{enumerate}
\subsection{Contribution}
\par This paper contributes to the field of automated content moderation by:
\begin{itemize}
    \item Providing a detailed analysis of ChatGPT's capabilities in detecting and classifying targeting and inappropriate language.
    \item Comparing the model's performance with crowd-sourced and expert annotations to gauge its effectiveness in real-world scenarios.
    \item Offering insights into iterative improvements and adjustments to refine the model's accuracy and reliability, including which adjustments did not help.
\end{itemize}
\subsection{Structure}
\par The paper is structured as follows: Section \ref{sec:literature} outlines the related work in content moderation and the application of NLP in this domain. Section \ref{sec:data} describes the data set employed in this study. Section \ref{sec:methodology} details the methodology employed to evaluate ChatGPT's performance in detecting targeting and inappropriate language within online comments. Section \ref{sec:results} presents the results from the evaluation of ChatGPT's performance in detecting targeting and inappropriate language, and discusses the implications of these findings as well as a more in-depth analysis of the expert annotations. Section \ref{sec:conclusion} concludes with implications for future research and potential applications of ChatGPT in automated content moderation systems, and the limitations of the study are set out thereafter.
\par By investigating ChatGPT's performance and iterating on its configurations, this study aims to advance the capabilities of automated content moderation tools and contribute to more effective and scalable solutions for managing online interactions.
\section{Literature Review}
\label{sec:literature}
\par This section provides a comprehensive review of existing literature on content moderation, with a focus on the evolution from manual to automated systems, the role of AI and NLP in enhancing moderation efforts, and the comparative effectiveness of different moderation strategies. By examining key studies and frameworks, we aim to contextualize our research within the broader landscape of content moderation and highlight the current state of knowledge and practice in this field.
\subsection{Overview of Content Moderation Challenges}
\par \cite{demetis2020breaking} discusses internet trolling and antisocial behavior on SNSs as significant challenges that disrupt online communities and contribute to a spectrum of dark and ultra-dark phenomena. The paper highlights how these behaviors manifest through various forms of exploitation, including identity deception, cyberbullying, and financial fraud, impacting both individuals and groups within SNSs. The prevalence of trolling and antisocial activities is noted for its widespread nature, exacerbated by the anonymity and reach of online platforms. The impact on online communities is profound, leading to heightened reputational risks, emotional distress, and the reinforcement of harmful behaviors through echo chambers and targeted exploitation. \cite{demetis2020breaking} underscores the need for comprehensive strategies to manage and mitigate these issues, involving proactive measures and stakeholder collaboration to protect users and enhance online safety.
\par \cite{gorwa2020algorithmic} note that human moderation faces significant limitations due to the sheer volume of online content, which is too vast for human moderators to effectively manage on their own. The immense scale of content being uploaded daily necessitates automated solutions to handle the massive influx efficiently. Automated systems, despite their imperfections, offer a scalable approach to content moderation by rapidly processing and flagging potentially problematic content. This need for automation is driven by the practical constraints of human capacity, which makes it essential to develop and refine technological solutions to supplement and enhance human efforts in maintaining content standards at scale.
\subsection{AI and NLP in Content Moderation}
\par \cite{gongane2022detection} provide a comprehensive overview of early AI and NLP methods for content moderation by detailing the evolution from manual to automated systems. Initially, social media platforms relied on human moderators to review and flag harmful content, but this approach faced scalability issues and inconsistencies. The introduction of semi-automated systems marked a shift towards using AI to flag content for human review, addressing volume challenges but introducing issues with bias and transparency. Early automated systems like PhotoDNA and ContentID used AI to filter content based on text, yet they struggled with multimedia content and global variations. \cite{gongane2022detection} highlight these early methods' limitations and emphasizes the need for further advancements in handling diverse content forms, multilingual challenges, and ensuring fairness and transparency in AI-driven moderation.
\par \cite{llanso2020no} examines how advancements in machine learning and deep learning, such as natural language processing and object recognition, have been applied to content moderation to enhance proactive filtering methods. They highlight that while these advanced AI techniques aim to improve the accuracy of detecting problematic content and reduce the limitations of simpler methods like keyword filtering and hash matching, they still face significant challenges. These challenges include the potential for overbreadth, inaccuracies, and the ease with which users can circumvent filters. Furthermore, \cite{llanso2020no} argues that regardless of the sophistication of these technologies, they do not resolve fundamental human rights concerns related to prior restraint and the lack of procedural safeguards in content moderation systems.
\par \cite{horta2021platform} discuss content moderation in the context of community-level interventions and their impacts by employing a regression discontinuity design and other statistical models. They utilize regression models to analyze changes in activity and content, including toxicity and radicalization signals, before and after community migration. Specifically, they mention adding a term for median post length to control for potential confounding effects related to changes in post formatting. Additionally, \cite{horta2021platform} reference frameworks such as the fixation dictionary for analyzing the prevalence of extremist content and LIWC (Linguistic Inquiry and Word Count) metrics for assessing emotional and behavioral signals, highlighting their use in evaluating the shifts in community content and user behavior.
\subsection{ChatGPT and Other Large Language Models}
\par \cite{roumeliotis2023chatgpt} detail ChatGPT's development by OpenAI as a sophisticated AI model built upon the GPT (Generative Pretrained Transformer) architecture, specifically an advanced variant of GPT-2. Its creation involved a two-phase process: an initial unsupervised pre-training phase on a vast corpus of text data, which enabled the model to learn language patterns and structures, followed by a supervised fine-tuning phase that refined its abilities on specific tasks like text completion and question-answering using labeled data sets. This extensive training process, which began with a prototype launch in November 2022 and public release in January 2023, has equipped ChatGPT with the capability to generate contextually coherent and human-like responses, making it a leading tool in natural language processing.
\par \cite{kumar2023watch} examine the use of large language models (LLMs) like ChatGPT in content moderation, highlighting their strengths in understanding complex human language and contextual nuances. They discuss how these models can be employed to filter and manage diverse and evolving online content by leveraging their advanced language comprehension capabilities. \cite{kumar2023watch} note that LLMs can adeptly interpret the subtleties of user-generated content, which enhances their effectiveness in moderating discussions, detecting harmful behavior, and ensuring adherence to community standards. Their dynamic adaptability allows them to address a wide range of content and adapt to new trends and issues in real-time.
\subsection{Comparative Studies with Human Moderation}
\par \cite{molina2024does} examine how individual differences influence perceptions of AI models versus human moderators in content classification systems. They find that users with low dispositional trust or conservative political views are more likely to view AI as accurate and objective compared to human moderators, which can enhance their overall trust in AI systems. Conversely, users who fear AI or have high technical expertise are more likely to perceive AI as limited in nuanced judgment, leading to reduced trust in AI compared to human moderation. This comparison underscores that while AI is perceived as more objective by certain users, others view it as lacking in subjective insight, highlighting the need for systems that transparently integrate both AI and human inputs to address these varied perceptions.
\par The \cite{bhattacharya2024towards} explore the role of crowd-sourced annotations in content moderation by using a diverse group of annotators to classify tweets from political discussions in India. They highlight that while crowd-sourcing can provide valuable insights and reflect a range of perspectives, achieving consistent and accurate annotations is challenging due to variability in annotators' judgments. This variability, or disagreement, is analyzed to understand how it might impact moderation decisions. The study finds that annotators often disagree on classifications, suggesting that content with high disagreement should be moderated more cautiously. This underscores the difficulty of relying on non-expert contributors for consistent content moderation, revealing the need for improved strategies and more robust training to enhance annotation reliability.
\subsection{Challenges and Limitations in Current Research}
\par \cite{lai2022human} highlights concerns related to bias in AI models by emphasizing how conditional delegation and explanations interact with model performance, particularly in content moderation. It reveals that while conditional delegation can improve rule precision when the AI model performs well (in-distribution), it may not adequately address model bias or performance issues when the AI encounters out-of-distribution data. The study also points out that global explanations, while intended to assist users, can inadvertently lead to biased rule creation due to their priming effect—participants might adopt high-frequency but less precise keywords without critically assessing their effectiveness. This suggests that while explanations can improve efficiency, they may also reinforce or introduce biases if not carefully designed, highlighting the need for strategies to mitigate such effects and better account for varying model performance and biases.
\par \cite{singh2020current} highlight several technical challenges that limit the practical application of AI models in real-world scenarios. These challenges include the complexity of integrating AI with existing systems, and the need for interoperability between new AI tools and current technologies. AI algorithms often operate as "black boxes," making it difficult to understand how they make decisions, which complicates their adoption. Additionally, the development and maintenance of in-house AI systems require significant resources for data infrastructure, labeling, and quality improvement. The paper also points out that initial AI products tend to have narrow applications and may lack broader utility, while the lack of established reimbursement models adds financial uncertainty. These technical hurdles collectively hinder the seamless integration and effective utilization of AI.
\par \cite{vidgen2020directions} highlight the significant challenges in accurately identifying abusive language due to several factors, including the inherent complexity and variability of online abuse, which spans a wide range of phenomena from identity-based hate to interpersonal abuse. Difficulties arise from the need for well-defined tasks and clear taxonomies, the reliance on limited and potentially biased sources of data, and the intricacies involved in annotating abusive content, such as dealing with irony and contextual nuances. Moreover, \cite{vidgen2020directions} note that many existing data sets suffer from poor documentation and limited methodological transparency, which exacerbates the issue by obscuring biases and making it harder to understand and address the limitations of detection systems. Consequently, these challenges necessitate careful data set creation, diverse and skilled annotators, and robust documentation to improve the accuracy and reliability of abusive language identification.
\subsection{Iterative Improvement of AI Models}
\par \cite{lo2023clear} focuses on improving AI model performance through the CLEAR Framework for Prompt Engineering, which provides a systematic approach to crafting effective prompts for AI language models. By emphasizing the components of Concise, Logical, Explicit, Adaptive, and Reflective prompts, the framework helps users design queries that are clear, well-structured, and tailored to elicit accurate and relevant responses from AI models. This structured approach ensures that prompts are optimized to leverage the full potential of AI algorithms, thereby enhancing the quality and coherence of the generated content and better meeting the specific needs of users.
\par \cite{lee2020clinical} describe how continual learning models enhance their performance by incrementally incorporating new data and feedback while retaining previously learned knowledge. These models dynamically adapt based on new information, allowing them to refine their predictions and tasks over time. This process mirrors human learning, where past experiences inform future decisions, and contrasts with static models that do not update after initial training.
\par In summary, the literature underscores the complex and evolving nature of content moderation in the digital age. From the limitations of human moderation due to scale and volume to the advancements in AI and NLP that offer scalable solutions, the field is marked by both significant progress and ongoing challenges. The reviewed studies reveal the strengths and shortcomings of various approaches, including early automated systems, sophisticated AI models like ChatGPT, and crowd-sourced annotations. As we continue to refine and enhance content moderation techniques, it is evident that a multi-faceted approach combining technological innovation with human oversight and continuous improvement is essential for addressing the diverse and dynamic challenges of online content management.
\section{Data}
\label{sec:data}
\begin{table}[ht]
\centering
\begin{tabular}{lccc}
\toprule
\textbf{Category} & \textbf{Crowd} & \textbf{Experts} & \textbf{ChatGPT} \\
\midrule
Targeting & 141 & 136 & 188 \\
Non-Targeting & 107 & 112 & 60 \\
\midrule
Inappropriate & 83 & 103 & 83 \\
Appropriate & 126 & 106 & 126 \\
\bottomrule
\end{tabular}
\caption{Annotation Results for Inappropriateness and Targeting Language in the Gold Set by Crowd, Experts, and ChatGPT. The targeting rows are computed over the 248 annotated units of the gold set, comprising 209 comments and 39 subthread titles, since titles were annotated for targeting; the inappropriateness rows are computed over the 209 comments alone. The identical inappropriateness totals for Crowd and ChatGPT are coincidental: the two sources agree on 179 of the 209 comments. The ChatGPT targeting figures correspond to the original prompt configuration, whose confusion matrix in Figure \ref{fig:chatgpt_original_version} gives $127 + 61 = 188$ units labelled targeting and $51 + 9 = 60$ labelled not targeting.}
\label{tab:data}
\end{table}

\begin{table}[ht]
\centering
\begin{tabular}{lcc}
\toprule
\textbf{Category} & \textbf{Crowd} & \textbf{ChatGPT} \\
\midrule
Targeting & 1689 & 1986 \\
Non-Targeting & 948 & 651 \\
\midrule
Inappropriate & 978 & 846 \\
Appropriate & 1167 & 1299 \\
\bottomrule
\end{tabular}
\caption{Annotation Results for Inappropriateness and Targeting Language in the Entire Data Set by Crowd and ChatGPT. The targeting rows are computed over 2,637 annotated units and the inappropriateness rows over 2,145 comments. The ChatGPT targeting figure is derived from the target-type annotation, counting any assigned target type as targeting (1,898 \textit{other target}, 68 \textit{another person in the conversation}, and 20 \textit{speaker}); the binary targeting-or-not annotation of the refined Version 6 configuration evaluated in Section \ref{sec:results} yields a different distribution over the same units.}
\label{tab:alldata}
\end{table}
\par We utilized an already existing data set of English conversation threads created by \citet{barbarestani2024content} from 28 banned subreddits on Reddit, accessed via the Pushshift API and BigQuery, comprising 67,677 submissions, 1,168,546 comments, and 4,017,460 tokens. These subreddits were chosen for their diverse representation of toxic behavior. Subthreads (conversation threads) were then filtered based on comments (3 to 17 per subthread), tokens (51 to 1,276 per subthread and up to 38 per comment), and toxicity. To determine toxicity, toxicity scores were calculated for each subthread by measuring the proportion of toxic words to all tokens. This calculation utilized three different lexicons: Wiegand by \citet{wiegand2018inducing}, Hurtlex by \citet{bassignana2018hurtlex}, and a lexicon developed by \citet{schouten2023cross} following the methodology presented by \citet{zhu2021hitsz}. Subthreads were then divided into 10 bins according to their normalized toxicity scores, which ranged from 0.08 to 0.2. Given that the majority of comments and subthreads on Reddit do not contain toxic words, a random selection of subthreads is likely to include many non-toxic entries. Therefore, to ensure a focus on toxic behavior, 400 subthreads were selected by \citet{barbarestani2024content} from the higher toxicity bins and an additional 98 subthreads with a toxicity score of 0. From the total 498 selected subthreads, including 400 high-toxicity subthreads (1,778 comments) and 98 zero-toxicity subthreads (367 comments), 39 were chosen as the gold data subset through stratified sampling. This sampling was based on toxicity scores and the number of comments, with a higher sampling rate for subthreads containing more than 6 comments. Communities such as CringeAnarchy and The\_Donald were highly represented. The gold subset comprises 209 comments, and 248 annotated units once the 39 subthread titles are included; these two figures are the denominators of the gold-set analyses reported throughout this paper.
\par According to \citet{barbarestani2024content}, comments were annotated for inappropriateness and targeting language, considering prior context. Annotators reviewed the comments and associated the context, including the title and previous comments within that conversation. For targeting language, titles were annotated as well as comments, which is why the two tasks have different denominators. For inappropriateness, the focus was on explicitly offensive language, including swear words, slurs, and profanity. For targeting language, annotators identified instances where comments or titles targeted individuals or groups based on categories such as sexual orientation, gender, disability, age, race/ethnicity/nationality, religion, political affiliation, and fame. Annotators focused primarily on nouns. The usernames of Reddit posters were anonymized, adhering to guidelines for handling potentially offensive content. Both annotation tasks were facilitated using the LingoTURK platform \citep{pusse2016lingoturk} and included annotators from Prolific \citep{palan2018prolific}. Annotations were conducted by experts (for the gold set), crowd annotators (for the entire data set), and ChatGPT (for the entire data set). The expert and crowd annotations were adjudicated based on the majority vote, and these adjudicated annotations are used in this study. Table \ref{tab:data} shows the annotation results for inappropriateness and targeting in the gold set, while Table \ref{tab:alldata} presents the annotation results across all data. The comparison among crowd annotators, experts, and ChatGPT provides insights into the consistency and reliability of these annotation sources in detecting inappropriate and targeting language. This helps in evaluating the potential use of automated systems like ChatGPT for annotation tasks.
\par The reliability figures reported by \citet{barbarestani2024content} distinguish three comparisons, and it is worth keeping them apart, since they are the reference points against which ChatGPT is assessed below. Among annotators of the same kind, the three expert annotators reached a Cohen's Kappa of 0.58 for targeting language at the comment level and 0.76 for inappropriateness at the token level, while the five crowd annotators reached 0.36 for targeting language at the comment level and 0.76 for inappropriateness at the token level. Between the two groups, adjudicated crowd and adjudicated expert annotations agree at 92\% percentage agreement for inappropriateness at the comment level. Between ChatGPT and the experts, agreement was moderate, with Kappa scores of 0.4 for targeting language and 0.88 for inappropriateness at the comment level. Two patterns in these figures shape the analyses that follow. First, targeting is consistently the harder judgment, for human annotators as much as for the model: no comparison exceeds 0.58 on targeting, whereas inappropriateness is annotated consistently in every comparison. Second, agreement on inappropriateness is high at the token level and higher still at the comment level, which indicates that annotators differ over which specific words are inappropriate more often than over whether a comment contains inappropriate language at all.
\section{Methodology}
\label{sec:methodology}
\par The goal of this study is to evaluate and enhance the performance of ChatGPT in identifying and classifying inappropriate and targeting language within online conversation threads. As online platforms increasingly rely on automated systems to manage content moderation at scale, it is crucial to assess how well these systems can replicate and improve upon human judgment. To achieve this, we have employed a multi-faceted methodology involving detailed analysis of conversation threads, annotation of content, and iterative refinement of ChatGPT's prompts and configurations. This section outlines our approach, including the segmentation of conversation threads to analyze the distribution of problematic behaviors, the calculation of annotation scopes to identify patterns in user comments, and the evaluation of ChatGPT's performance under various prompting scenarios. By systematically testing and improving the model's ability to detect harmful content, we aim to provide insights into its effectiveness and identify strategies for enhancing its accuracy in real-world applications.
\par We primarily compared ChatGPT annotations to the crowd annotations on the gold set. Crowd annotations provide a diverse range of perspectives that reflect the variability of real-world content moderation environments. The agreement between crowd and expert annotations on the gold data was 92\% for inappropriate language and 58\% for targeting language. These figures indicate that crowd annotations align closely with expert judgments for inappropriate language and moderately so for targeting language, which is itself the harder of the two judgments throughout this study. Using crowd annotations as the benchmark allowed us to evaluate how well ChatGPT's performance aligns with a substantial amount of content moderation scenarios, providing a robust validation method. The decision to primarily analyze the gold set, rather than the entire data set, was driven by the need to focus on the most reliable and representative samples for evaluating targeting and inappropriateness detection.
\subsection{Analysis of Conversation Threads}
\par To better understand the distribution and manifestation of inappropriate and targeting behaviors identified in online conversations, we segment conversation threads into distinct stages for detailed analysis. The primary goal is to identify how these behaviors appear at different points in a conversation and whether their prevalence varies throughout the interaction. To achieve this, conversation threads from both crowd and ChatGPT annotations were divided into three bins: beginning, middle, and final parts. For inappropriate content, which excludes the title text at the start, the bins were defined as follows: the beginning comprised the first 34\% of the conversation, the middle ranged from 34\% to 67\%, and the final covered from 67\% to the end. For targeting behavior, the title text is included as the first unit, and the bins were adjusted accordingly: the beginning up to 33\%, the middle from 33\% to 66\%, and the final from 66\% to the end. The cut-points therefore differ slightly between the two behaviors because the two analyses count a different number of units per thread. The results are detailed in Tables \ref{tab:onset_targeting}, \ref{tab:prevalence_targeting}, \ref{tab:onset_inappropriateness}, and \ref{tab:prevalence_inappropriateness}, enabling a comprehensive examination of how the evolution of inappropriate and targeting behaviors is identified throughout a conversation. Throughout these tables, percentages are computed over all 39 gold-set subthreads, so the onset columns do not sum to 100\%.
\subsection{Annotation Scopes}
\par We calculated the average number of comments per conversation thread, forming a scope of consecutively annotated comments as inappropriate or targeting within each annotation set (see the results in Table \ref{tab:average_annotation_scopes}). Title text annotations were also included in the targeting analysis. These calculations help to reveal the identification of patterns of sustained engagement in such behaviors, which are critical for understanding the dynamics of harmful interactions. The insights gained can inform the development of targeted moderation strategies aimed at reducing the prevalence of inappropriate or targeting behaviors.
\subsection{The Role of Contextual Cues in Detecting Targeting and Inappropriate Language by ChatGPT}
\label{sec:chatgpt_prompt}
\par We first evaluate ChatGPT's capability to identify inappropriate and targeting behavior using the originally adjudicated expert labels from the data set. The purpose of this evaluation was to assess how well ChatGPT could detect these behaviors when provided with contextual cues, based on its initial understanding (see Table \ref{tab:chatgpt_results}). To achieve this, we developed a set of prompts designed to test ChatGPT's ability to interpret contextual information within conversation threads. We aim to investigate ChatGPT's performance when provided with relevant details about the conversation, including preceding comments (context) and their labels, which we call `contextual cues'. ChatGPT was then asked to determine if the comment exhibited inappropriateness and/or targeting. The evaluation was conducted under the following scenarios:
\begin{enumerate}
    \item \label{enum:scenario1} \textbf{Scenario 1:} ChatGPT received targeting labels from the context, including previous comments and the targeting label for the comment being processed. It then determined if this comment was inappropriate. Note that the comment was targeting, but this was unknown to ChatGPT. This comment was the first inappropriate one following a targeting comment in the conversation thread.
    \item \label{enum:scenario2} \textbf{Scenario 2:} Without providing targeting labels from previous comments or for the comment under consideration, ChatGPT was asked to determine if the comment exhibited inappropriateness. Note that the comment was targeting, but this was unknown to ChatGPT. This comment was the first inappropriate one following a targeting comment in the conversation thread. The comment selected is the same as in Scenario \ref{enum:scenario1}, so the two scenarios differ only in the information supplied.
    \item \label{enum:scenario3} \textbf{Scenario 3:} Without context labels for targeting and inappropriateness, as well as no such labels for the comment under consideration, ChatGPT had to determine if the target comment exhibited either behavior. The comment was both targeting and inappropriate, which was unknown to ChatGPT, and was the first such comment in the conversation thread.
    \item \label{enum:no labels context last} \textbf{Scenario 4:} Without labels for targeting and inappropriateness in the context or the target comment, ChatGPT was asked to determine if the target comment exhibited targeting or inappropriateness. The processed comment was neither targeting nor inappropriate, but this was unknown to ChatGPT. This comment was the last in the conversation thread and was neither targeting nor inappropriate.
\end{enumerate}
\par Each scenario contributes one comment from every gold-set subthread that satisfies its condition, so the four scenarios rest on subsets of different size: 14 subthreads for Scenarios \ref{enum:scenario1} and \ref{enum:scenario2}, 17 for Scenario \ref{enum:scenario3}, and 29 for Scenario \ref{enum:no labels context last}. Scenarios \ref{enum:scenario1} and \ref{enum:scenario2} are drawn from the same 14 subthreads and are therefore directly comparable with one another; comparisons across the remaining scenarios are not, since they concern different comments. Given these subset sizes, the agreement figures reported in Section \ref{sec:results} are indicative rather than precise, and we treat only differences substantially larger than a single unit as informative. In scenarios \ref{enum:scenario1}, \ref{enum:scenario2}, and \ref{enum:scenario3}, ChatGPT's responses were based on initial encounters with potentially inappropriate or targeting comments, simulating real-time moderation. We utilized the GPT-4 model for this evaluation. The prompts for each scenario can be found in Appendix \ref{appendix:chatgpt_scenarios}.
\subsection{Prompt Improvement Process for the Identification of Targeting Language}
\begin{figure}[h!]
    \centering
    \includegraphics[width=\textwidth]{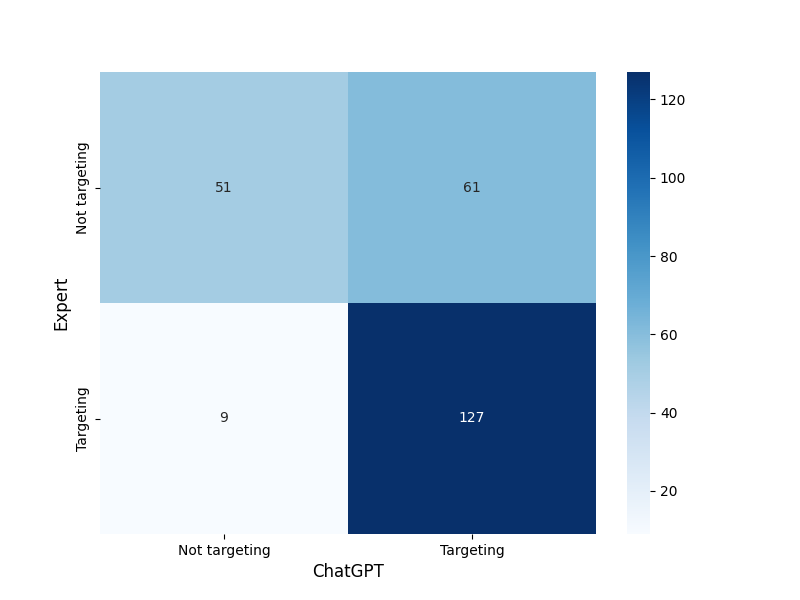}
    \vspace{10pt}
    \caption{Comparison of the Performance of ChatGPT Original Version in Targeting Language Identification to the Expert Set}
    \label{fig:chatgpt_original_version}
\end{figure}
\par Here, we discuss a separate analysis from Subsection \ref{sec:chatgpt_prompt} that compares the newly generated targeting and inappropriate labels by ChatGPT with the original expert annotations in the gold set. The initial comparison between ChatGPT's original annotations and the expert annotations yielded a Cohen's Kappa score of 0.4, indicating only moderate agreement. Specifically, as shown in Figure \ref{fig:chatgpt_original_version}, there were 9 comments labeled as targeting by experts but not by ChatGPT, 127 comments consistently labeled as targeting by both, 51 comments consistently labeled as not targeting by both, and 61 comments where ChatGPT labeled them as targeting while experts did not. These counts run over the 248 annotated units of the gold set --- 209 comments and 39 subthread titles --- since targeting was annotated on titles as well as comments. For brevity they are referred to as comments in the discussion of the individual versions below. This moderate level of agreement suggested that ChatGPT's original prompts did not fully capture the nuances of targeting language as recognized by human annotators. To address this, we focused on refining the original prompts to enhance ChatGPT's ability to identify targeting language more accurately. This involved developing and testing six different prompt versions. Detailed descriptions of each prompt version and code, including the specific changes made, are provided in Appendix \ref{appendix:chatgpt_improvements}. This approach allowed us to identify specific areas where ChatGPT's annotations diverged from expert assessments, enabling targeted improvements. After validating these refinements with the expert set, we tested the improved prompts on a larger data set with crowd annotations, ensuring that the refinements were both accurate and broadly applicable. The crowd annotations, applied to a larger data set compared to the expert annotations, allowed for extensive validation of ChatGPT's performance. As mentioned earlier, the crowd annotations are sufficiently similar to those from the expert set to facilitate large-scale evaluation. By first using the expert set to fine-tune the prompts, we could effectively address the nuances of targeting language as identified by domain experts. Subsequently, the crowd evaluation captures a wider set of data that is more representative.
\begin{enumerate}
    \item \label{enum:version1} \textbf{Version 1:} Simplified the categorization of targeting language into a binary distinction (``TARGETING'' vs.\ ``NOT TARGETING'') and switched to the ``gpt-4'' model for improved clarity and consistency.
    \item \label{enum:version2} \textbf{Version 2:} Rewrote the instructions as numbered steps, asked the model to respond only with one of the two labels, and introduced a modular structure for prompt management with a retry mechanism.
    \item \label{enum:version3} \textbf{Version 3:} Rewrote the instructions again, asking whether the text clearly targets or attacks a specific person or group, and added comprehensive logging and refined error handling.
    \item \label{enum:version4} \textbf{Version 4:} Transitioned to the ``gpt-4o'' model while maintaining the improvements from Version \ref{enum:version3}.
    \item \label{enum:version5} \textbf{Version 5:} Reverted to ``gpt-4'' and adjusted the temperature parameter to 0.3 for more deterministic responses, retaining the improvements from previous versions.
    \item \label{enum:version6} \textbf{Version 6:} Further adjusted the temperature setting from 0.3 to 0 for even more consistent and less varied responses, while maintaining the enhancements of Version \ref{enum:version5}.
\end{enumerate}
\par Two kinds of change are combined in this sequence, and separating them matters for interpreting the results. Versions \ref{enum:version1} through \ref{enum:version3} revise the instructions given to the model, whereas Versions \ref{enum:version4} through \ref{enum:version6} share a single prompt and differ only in the model and the sampling temperature. Changes in agreement within the latter group are therefore attributable to the model and its sampling behavior rather than to the task specification, and the logging and error-handling refinements introduced in Versions \ref{enum:version2} and \ref{enum:version3} cannot themselves alter the labels the model returns.
\section{Results and Discussion}
\label{sec:results}
\subsection{Gold Set Analysis}
\subsubsection{Targeting Analysis}
\par We examined the onset and prevalence of targeting behavior within 39 conversation threads from the gold set, using annotations provided by the crowd and ChatGPT. Our analysis uncovered distinct patterns discernible across these annotation sources.
\paragraph{Onset of Targeting}
\begin{table}[h]
    \centering
    \begin{tabular}{lcc}
        \toprule
        \textbf{Segment} & \textbf{Crowd} & \textbf{ChatGPT} \\
        \midrule
        Beginning & 46\% & 38\% \\
        Middle & 41\% & 59\% \\
        Final & 0\% & 0\% \\
        \bottomrule
    \end{tabular}
    \caption{Onset of Targeting. Percentages are computed over all 39 gold-set subthreads and therefore do not sum to 100\%.}
    \label{tab:onset_targeting}
\end{table}

\par Table \ref{tab:onset_targeting} shows that crowd workers identify the early stages of conversations as critical for targeting initiation, with 46\% of subthreads beginning with targeting. ChatGPT identifies slightly fewer instances at the beginning (38\%). In the middle stages, crowd workers identify targeting in 41\% of subthreads, while ChatGPT detects more such instances, at 59\%. Importantly, neither the crowd nor ChatGPT annotations record a single subthread in which targeting begins in the final stage. Targeting behaviors therefore emerge in the early and middle stages of conversations, underscoring the importance of early detection.
\begin{table}[htbp]
\centering
\small
\begin{tabular}{lccc}
\toprule
\textbf{Token} & \textbf{Frequency} & \textbf{POS} & \textbf{Token Type} \\
\midrule
stupid & 9 & JJ & offensive \\
gay & 8 & NN, JJ & non-offensive \\
f**king & 8 & JJ, VBG & offensive \\
the & 8 & DT & non-offensive \\
\bottomrule
\end{tabular}
\caption{Tokens Marking the Onset of Targeting in Conversation Threads. Onset tokens may stand alone or form part of a longer annotated span, which is why function words such as \textit{the} appear among the most frequent.}
\label{tab:analysis}
\end{table}
\par Additionally, we extracted the frequencies, token types, and part of speech (POS) tags of the tokens marking the onset of targeting, which could belong to longer spans or be single tokens on their own, in all subthreads and comments from the entire data set annotated by the crowd as ``targeting''. This token analysis is the one part of this subsection that ranges beyond the gold set. The purpose of the analysis is to identify and understand the linguistic markers of targeting behavior in conversation threads. The token types were annotated by the authors of this paper. Table \ref{tab:analysis} illustrates the tokens with the highest frequency, all variants of their POS tags in the data, and token types. The analysis reveals a significant presence of offensive tokens (22) and slurs (16), including tokens like ``stupid,'' ``f**king,'' ``f**k,'' ``c**t,'' ``d**k,'' and various racial and homophobic slurs. Despite this, the majority of the tokens (223 tokens) are non-offensive. Note that a token can name a targeted group without itself being derogatory, which is why ``gay'' is classified as non-offensive here while ``f**king'' is not. The part-of-speech distribution is diverse, with nouns (NN: 39.49\%) and adjectives (JJ: 18.18\%) being the most frequent tags.

\paragraph{Prevalence of Targeting}
\begin{table}[ht]
\centering
\begin{tabular}{lcc}
\toprule
\textbf{Segment} & \textbf{Crowd} & \textbf{ChatGPT} \\
\midrule
Beginning & 46\% & 97\% \\
Middle & 85\% & 85\% \\
Final & 77\% & 85\% \\
\bottomrule
\end{tabular}
\caption{Prevalence of Targeting. Percentages are the share of all 39 gold-set subthreads in which targeting was annotated within the given segment; a subthread may contribute to more than one segment.}
\label{tab:prevalence_targeting}
\end{table}

\par The prevalence of targeting behavior across conversation thread segments is analyzed in Table \ref{tab:prevalence_targeting}. Crowd workers identify an increase in targeting from the beginning to the middle of conversations, followed by a slight decline in the final part, with 46\%, 85\%, and 77\% of subthreads respectively. In contrast, ChatGPT detects a higher prevalence of targeting at the beginning (97\%) and maintains a consistent detection rate of 85\% in both the middle and final parts of the conversations. The higher detection rate by ChatGPT at the beginning may indicate high sensitivity to early signs of targeting. This heightened sensitivity could potentially lead to over-detection and an increase in false positives. While human annotators perceive targeting as escalating or persisting as conversations progress, ChatGPT's consistent detection across the conversation parts highlights its different sensitivity profile.

\paragraph{Development of Targeting}
\begin{table}[ht]
\centering
\begin{tabular}{lcc}
\toprule
\textbf{Targeting Development} & \textbf{Crowd} & \textbf{ChatGPT} \\
\midrule
Targeting starts based on an attribute & 59\% & 67\% \\
Targeting starts not based on an attribute & 0\% & 0\% \\
Attributed targeting becomes specific to an individual & 23\% & 28\% \\
\bottomrule
\end{tabular}
\caption{Distribution of Targeting Development in Conversation Threads Based on Attributes and Personalization. The third row describes a progression within the cases counted in the first row and is therefore a subset of it, not a further category.}
\label{tab:targeting_development}
\end{table}

\par Table \ref{tab:targeting_development} illustrates how targeting evolves in conversations based on identifiable attributes such as religion, political affiliation, race, sexual orientation, gender, disability, age, fame, and other factors. This analysis is crucial for understanding the dynamics of targeting behavior in online conversations. The data show the distribution of targeting behaviors identified across gold subthreads starting from general attributes to more personalized attacks. Targeting Based on an Attribute refers to targeting behaviors linked to specific personal characteristics. For instance, a comment like ``Incels f**king b***hes? Nah...'' targets individuals based on gender. In this case, the targeting is attributed to the gender of the individuals being discussed. This type of targeting was identified as the starting point of targeting in 59\% of cases by crowd annotators and 67\% by ChatGPT. Targeting Not Based on an Attribute describes instances where targeting does not involve specific characteristics shared with a group but focuses on general or contextual criticisms. For example, the comment ``I made a simple mistake and you call me idiotic 10/10 fu**kin logic mate'' targets someone based on the perceived logic of their statement rather than any personal attribute. Neither crowd annotators nor ChatGPT identified any cases of this type of targeting as the starting point of targeting in the data set. Attributed Targeting Becomes Specific to an Individual represents a progression from attribute-based targeting as the starting point to more personalized attacks. Initially, targeting may be based on an attribute shared with a group (e.g., ``go back to your f**king estro weed subs my dude. your rotten brain and sh*t comments belong with the other addicts''), but it can evolve into more specific targeting directed towards an individual (e.g., ``I mean the individual words appear to be English, most of them anyway, but I can't make out what this ghetto speak trash is even attempting to say..''). This type of personalized targeting was observed in 23\% of cases by crowd annotators and 28\% by ChatGPT. The two annotation sources order these developments in the same way, which is consistent with ChatGPT recognizing the broad shape of targeting dynamics, although the gold set is too small for the differences between the two columns to be interpreted. Distinguishing between targeting that starts from an attribute, targeting that starts without one, and the progression to personalized targeting highlights the need for early intervention strategies to mitigate the impact of such targeting in online conversations.

\subsubsection{Inappropriateness Analysis}
\par We examined the initiation and distribution of inappropriateness within a total of 39 conversation threads, using annotations from both the crowd and ChatGPT.
\paragraph{Onset of Inappropriateness}
\begin{table}[ht]
\centering
\begin{tabular}{lcc}
\toprule
\textbf{Segment} & \textbf{Crowd} & \textbf{ChatGPT} \\
\midrule
Beginning & 56\% & 64\% \\
Middle & 26\% & 20.5\% \\
Final & 0\% & 0\% \\
\bottomrule
\end{tabular}
\caption{Onset of Inappropriateness. Percentages are computed over all 39 gold-set subthreads and therefore do not sum to 100\%.}
\label{tab:onset_inappropriateness}
\end{table}

\par The data in Table \ref{tab:onset_inappropriateness} show that inappropriateness predominantly emerges at the beginning of conversations for both the crowd (56\% of subthreads) and ChatGPT (64\% of subthreads), indicating that initial interactions set the tone for subsequent discourse. This trend suggests that early exchanges have a significant impact on conversation dynamics, possibly due to users initially feeling less constrained by social norms or cues. The onset figures decline across the three segments, but this decline is partly definitional, since a subthread whose inappropriateness begins early cannot also have its onset counted later. What the figures establish is that inappropriateness, where it occurs at all, is almost always introduced in the first third of the exchange rather than developing later.

\paragraph{Prevalence of Inappropriateness}
\begin{table}[ht]
\centering
\begin{tabular}{lcc}
\toprule
\textbf{Segment} & \textbf{Crowd} & \textbf{ChatGPT} \\
\midrule
Beginning & 56\% & 64\% \\
Middle & 59\% & 54\% \\
Final & 56\% & 54\% \\
\bottomrule
\end{tabular}
\caption{Prevalence of Inappropriateness. Percentages are the share of all 39 gold-set subthreads in which inappropriateness was annotated within the given segment. The beginning row coincides with that of Table \ref{tab:onset_inappropriateness} by construction, since a subthread with inappropriateness in the first segment necessarily has its onset there.}
\label{tab:prevalence_inappropriateness}
\end{table}

\par Table \ref{tab:prevalence_inappropriateness} shows that inappropriateness is consistently present across different segments of the conversation. Specifically, inappropriate behavior is noted in 56\% of subthreads at the beginning, 59\% in the middle, and 56\% at the end of conversations by the crowd, and 64\%, 54\%, and 54\% respectively by ChatGPT. These data indicate that once inappropriate behavior is introduced, it tends to persist throughout the conversation. The differences between segments are small in both annotation sources and no escalation is inferred from them. This stability underscores the pervasive nature of toxic interactions in online environments and highlights the need for targeted moderation strategies: even towards the end of conversations, inappropriate behavior remains a significant issue, impacting the overall quality and tone of online discussions.

\subsubsection{Initial Detection of Targeting with Respect to Inappropriateness}
\begin{table}[h]
\centering
\begin{tabular}{lcc}
\toprule
\textbf{Detection Time of Inappropriateness vs. Targeting} & \textbf{Crowd} & \textbf{ChatGPT} \\
\midrule
Inappropriateness before targeting & 2.6\% & 49\% \\
Inappropriateness at the same time as targeting & 36\% & 23\% \\
Targeting before inappropriateness & 43.6\% & 13\% \\
\bottomrule
\end{tabular}
\caption{Analysis of the Detection Time of Inappropriateness in Relation to Targeting in the Conversation Threads. Percentages are computed over all 39 gold-set subthreads and therefore do not sum to 100\%.}
\label{tab:inappropriateness_targeting}
\end{table}

\par Table \ref{tab:inappropriateness_targeting} presents the timing of inappropriateness detection relative to targeting detection in conversation threads, based on annotations by the crowd and ChatGPT. The crowd's annotations reveal that inappropriateness is most often identified either simultaneously with targeting or after it, with inappropriateness preceding targeting in only 2.6\% of subthreads. ChatGPT reverses this ordering, identifying inappropriateness before targeting in 49\% of subthreads. This is the largest divergence between the two annotation sources reported here, and it suggests that the model registers surface offensiveness earlier than it registers directedness, which is consistent with the over-detection of targeting noted above. Overall, while both the crowd and ChatGPT detect targeting and inappropriateness, their patterns of detection indicate differing sensitivities and priorities in recognizing these behaviors within conversation threads.

\subsubsection{Annotation Scopes}
\par Based on the findings in Table \ref{tab:average_annotation_scopes}, we observe notable variations in identifying targeting and inappropriate language between different annotation methods. ChatGPT identified, on average, the broadest scopes of targeting language (4.13 comments), followed by the crowd (2.75 comments). For inappropriate language the two sources are close, with ChatGPT marginally longer (2.156 comments) than the crowd (2.08 comments). The difference is therefore specific to targeting: ChatGPT extends targeting over substantially more consecutive comments than crowd annotators do, which may also be attributed to a high false positive detection rate, whereas for inappropriateness the two sources agree on scope length.
\begin{table}[h]
    \centering
    \begin{tabular}{lcc}
        \toprule
        \textbf{Category} & \textbf{Crowd} & \textbf{ChatGPT} \\
        \midrule
        Targeting & 2.75 & 4.13 \\
        Inappropriate & 2.08 & 2.156 \\
        \bottomrule
    \end{tabular}
    \caption{Average Number of Comments Forming a Scope Representing Targeting and Inappropriateness}
    \label{tab:average_annotation_scopes}
\end{table}

\subsubsection{Patterns in Expert Annotation of Targeting Language}
\par Additionally, we analyzed comment-level annotations in each subthread identified for targeting language by the experts and categorized nuanced patterns in the interactions among the interlocutors, where fluctuations were observed: 
\begin{enumerate}
    \item External Targeting Cascade: Targeting initiates with insults or attacks towards individuals outside the conversation, provoking internal attacks among participants.
    \begin{exm}
    \begin{description}
        \item 
        \item anon\_kh8gg: Post the picture of Donald Jr with his kids. Jesus christ hes an ugly son of a b***h - that's the cringe
    \end{description}
    
    \begin{description}
        \item anon\_tkaeo : go back to your f**king estro weed subs my dude. your rotten brain and sh*t comments belong with the other addicts
    \end{description}
    
    \begin{description}
        \item anon\_kh8gg: Lol freedom fighter. You're a r**n**k fa**ot bro foh
    \end{description}
    \label{exm:external targeting}
    \end{exm}
    \item Non-provoked External Targeting: A participant targets someone outside the conversation without provoking further reactions.
    \begin{exm}
    \begin{description}
        \item 
        \item anon\_Khiib: Why the hell do You Care if guys like d**k instead of p***y. Relax dude
    \end{description}
    
    \begin{description}
        \item anon\_4OsvY: They're degenerate f**ks.
    \end{description}
    
    \begin{description}
        \item anon\_BbKt2: Dann, how long it take you to write that thesis?
    \end{description}
    \label{exm:non-provoked external}
    \end{exm}
    \item Mutual External Targeting: Participants target individuals outside the conversation, reinforcing each other's hostile behavior.
    \begin{exm}
    \begin{description}
        \item 
        \item anon\_Ic8wr: You do realize most furries aren't this insane right?
    \end{description}
    
    \begin{description}
        \item anon\_T49PP: It is not okay to be a "furry". You either have some weird species dismorphia, or a weird species fetish. Neither are okay. 

    \end{description}
    
    \label{exm:mutual external}
    \end{exm}
    \item Internal Targeting with Interlocutor Mediation: Participants attack each other, but other interlocutors intervene to stop the provocation.
    \begin{exm}
    \begin{description}
        \item 
        \item anon\_GQt1h: >"EVERYTHING I DON'T LIKE IS RUSSIAN PROPAGANDA"  Go back to /r/politics 
    \end{description}
    
    \begin{description}
        \item anon\_srpgg: hey s**k my d**k you f**king q***r

    \end{description}
    \begin{description}
        \item anon\_GQt1h: F**k off fa**ot

    \end{description}
    \begin{description}
        \item anon\_hMcof: Please never change

    \end{description}
    \begin{description}
        \item anon\_GQt1h: >being this much of a retard  Imagine actually believing everything that is anti-EU is Russian propaganda. Please never change being a retard.

    \end{description}
    \begin{description}
        \item anon\_srpgg: shut up b***h

    \end{description}
    
    \label{exm:internal targeting}
    \end{exm}
    \item Internal Targeting with De-escalation by an Interlocutor: Internal targeting begins but is later de-escalated by one of the participants.
    \begin{exm}
    \begin{description}
        \item 
        \item anon\_3PKni: Only aid to Israel!  Please keep sending aid to Israel, a country who knows how to treat other races fairly!

    \end{description}
    
    \begin{description}
        \item anon\_MO4S0: AIDS to israel? You antisemite.

    \end{description}
    \begin{description}
        \item anon\_3PKni: Ya sjw c**k c**k c**kc**kc**kc**kc**kc**kc**k error error reset c**k c**k c**k c**k c**k c**k

    \end{description}
    \begin{description}
        \item anon\_MO4S0: Nevah foget da 6 billion jews that were gassed and turned into ashes

    \end{description}
    
    \label{exm:internal targeting with de-escalation}
    \end{exm}
\end{enumerate}
\subsubsection{Evaluating ChatGPT's Performance in Detecting Targeting and Inappropriate Language: The Role of Contextual Cues}
\begin{table}[h]
\centering
\begin{tabular}{lrcc}
\toprule
\textbf{Scenario} & \textbf{N} & \textbf{Agreement on Inappropriateness} & \textbf{Agreement on Targeting} \\
\midrule
1 & 14 & 71.43\% & - \\
2 & 14 & 64.29\% & - \\
3 & 17 & 94.12\% & 47.06\% \\
4 & 29 & 96.55\% & 79.31\% \\
\bottomrule
\end{tabular}
\caption{ChatGPT's Agreement on Inappropriateness and Targeting After Prompting Using Experts' Original Labels. $N$ is the number of subthreads contributing one comment each, determined by the selection condition of each scenario as defined in Subsection \ref{sec:chatgpt_prompt}. Scenarios \ref{enum:scenario1} and \ref{enum:scenario2} draw on the same 14 subthreads and are therefore directly comparable; Scenarios \ref{enum:scenario3} and \ref{enum:no labels context last} draw on different subthreads and are not comparable to them.}
\label{tab:chatgpt_results}
\end{table}

\par The observations from Table \ref{tab:chatgpt_results} highlight ChatGPT's varying levels of agreement with expert annotations on inappropriateness and targeting with and without contextual cues. The results indicate that ChatGPT performs better in identifying inappropriate comments than in detecting targeting behavior. The most direct comparison is internal to a single scenario. Scenario \ref{enum:scenario3} presents identical input for both judgments and yields 94.12\% agreement on inappropriateness against 47.06\% on targeting, that is, 16 of 17 units against 8 of 17; Scenario \ref{enum:no labels context last} shows the same ordering, 28 of 29 against 23 of 29. Because the available information is the same within each scenario, the gap cannot be attributed to differences in context. It is consistent with the different nature of the two judgments, although part of it may reflect the inverted label convention of the targeting question (Section \ref{sec:limit}): inappropriateness is substantially recoverable from the lexical properties of the comment, whereas targeting requires deciding whom the comment positions as its object, which the surface form underdetermines. Scenarios \ref{enum:scenario1} and \ref{enum:scenario2} address a different question, namely whether supplying the targeting labels of the preceding comments improves the identification of inappropriateness. Both scenarios were evaluated over the same 14 subthreads, and agreement was 71.43\% with those labels against 64.29\% without, a difference of one comment. We therefore do not treat this as evidence that contextual cues improve the identification of inappropriate content, and note only that the effect, if any, is smaller than this evaluation can resolve. The variability in agreement rates for targeting behavior underscores the need for further refinement in the model's ability to handle nuanced targeting scenarios.
\subsubsection{Prompt Improvement Results for the Identification of Targeting Language}

\begin{figure}[h!]
    \centering
    \includegraphics[width=\textwidth]{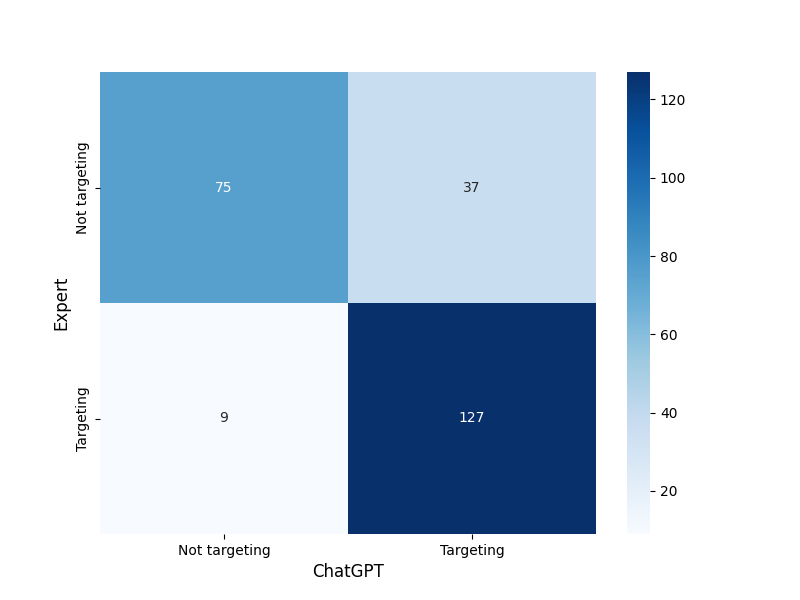}
    \vspace{10pt}
    \caption{Comparison of the Performance of ChatGPT Version 1 in Targeting Language Identification to the Expert Set}
    \label{fig:chatgpt_version1}
\end{figure}

\paragraph{Version \ref{enum:version1}} Comparing Version \ref{enum:version1} (Figure \ref{fig:chatgpt_version1}) to the original version (Figure \ref{fig:chatgpt_original_version}), notable improvements in performance and agreement with expert annotations are evident. Version \ref{enum:version1} achieved a Cohen's Kappa score of 0.62, a substantial increase from the original version's score of 0.4, indicating a significant enhancement in the agreement between ChatGPT's labeling and the expert annotations. This improvement is supported by a reduction in the number of comments classified inconsistently between ChatGPT and experts. Specifically, the original version had 61 instances where ChatGPT labeled a comment as ``targeting'' that experts did not, which Version \ref{enum:version1} reduced to 37. Similarly, the number of comments correctly classified as ``not targeting'' by both ChatGPT and experts increased from 51 to 75. The number of comments classified as ``targeting'' by both ChatGPT and experts remained the same at 127, so the increase in Cohen's Kappa is due entirely to better alignment on content that was ``not targeting''. The model did not become better at recognizing targeting; it became better at recognizing its absence. These improvements highlight the efficacy of simplifying the prompt and switching to a more advanced model, contributing to a clearer and more accurate identification of targeting language and bridging the gap between automated and expert assessments.

\begin{figure}[h!]
    \centering
    \includegraphics[width=\textwidth]{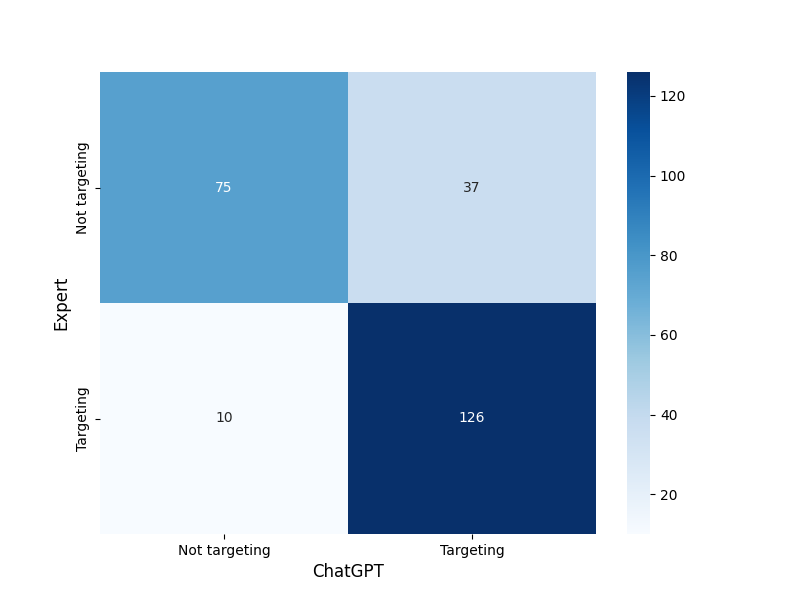}
    \vspace{10pt}
    \caption{Comparison of the Performance of ChatGPT Version 2 in Targeting Language Identification to the Expert Set}
    \label{fig:chatgpt_version2}
\end{figure}

\paragraph{Version \ref{enum:version2}} Analyzing Version \ref{enum:version2} compared to Version \ref{enum:version1} reveals a slight decrease in Cohen's Kappa score from 0.62 to 0.61, indicating a marginal reduction in agreement with expert annotations. As shown in Figure \ref{fig:chatgpt_version2}, Version \ref{enum:version2} slightly reduced the count of ``targeting'' comments agreed upon by both ChatGPT and experts from 127 to 126. The number of consistently identified ``not targeting'' comments remained the same at 75 for both versions. However, the number of ``targeting'' comments identified by experts but missed by ChatGPT increased from 9 in Version \ref{enum:version1} to 10 in Version \ref{enum:version2}. Similarly, the number of ``not targeting'' comments incorrectly identified as ``targeting'' by ChatGPT remained unchanged at 37 in both versions. These results suggest that the rewritten instructions of Version \ref{enum:version2} together with its improvements in error handling and code organization, did not enhance agreement with expert labels compared to Version \ref{enum:version1}.

\begin{figure}[h!]
    \centering
    \includegraphics[width=\textwidth]{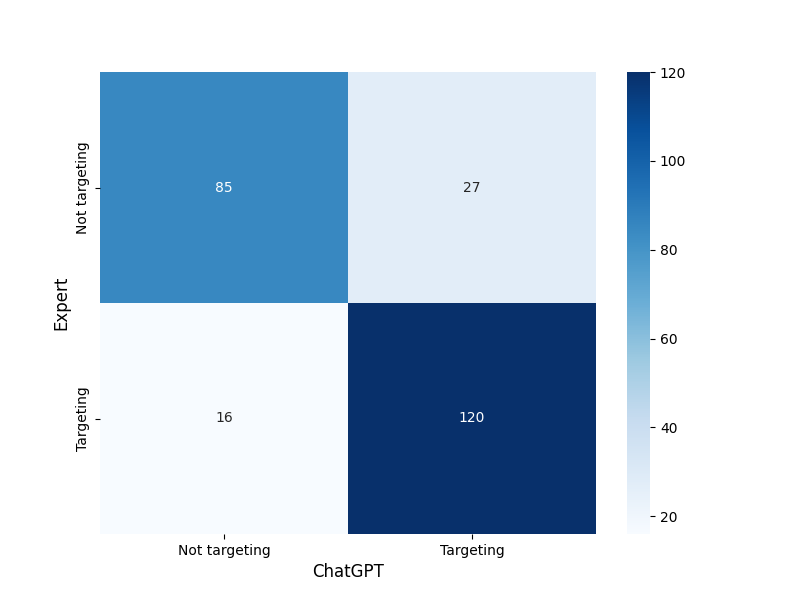}
    \vspace{10pt}
    \caption{Comparison of the Performance of ChatGPT Version 3 in Targeting Language Identification to the Expert Set}
    \label{fig:chatgpt_version3}
\end{figure}

\paragraph{Version \ref{enum:version3}} Version \ref{enum:version3} shows an improvement in agreement with human annotations compared to Version \ref{enum:version2}, with a higher Cohen's Kappa score of 0.65, up from 0.61. As seen in Figure \ref{fig:chatgpt_version3}, Version \ref{enum:version3} correctly identified 120 ``targeting'' comments in agreement with experts, compared to 126 in Version \ref{enum:version2}. It showed more false negatives (16 comments where experts identified ``targeting'' but ChatGPT did not, compared to 10 in Version \ref{enum:version2}) and fewer false positives (27 instances of ChatGPT identifying ``targeting'' where experts did not, compared to 37 in Version \ref{enum:version2}). Additionally, Version \ref{enum:version3} had a higher number of consistent ``not targeting'' classifications (85 vs.\ 75 in Version \ref{enum:version2}). The gain in Kappa comes from the revised prompt trading recall for precision, rather than from the logging and error-handling changes introduced in this version, which do not affect the model's output. The same trade-off is visible in the counts: the model marks fewer comments as targeting overall, and is more often right when it does.

\begin{figure}[h!]
    \centering
    \includegraphics[width=\textwidth]{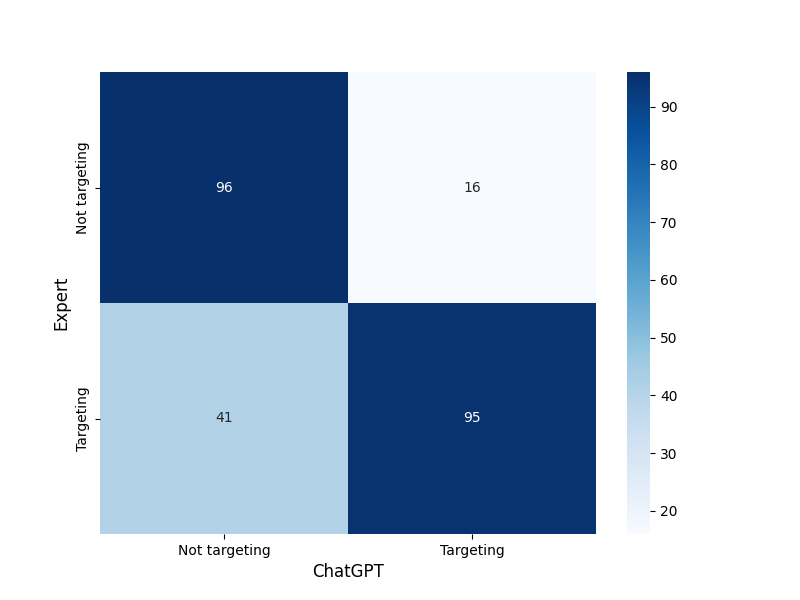}
    \vspace{10pt}
    \caption{Comparison of the Performance of ChatGPT Version 4 in Targeting Language Identification to the Expert Set}
    \label{fig:chatgpt_version4}
\end{figure}

\paragraph{Version \ref{enum:version4}} The results of Version \ref{enum:version4} show a Cohen's Kappa score of 0.54, indicating a moderate level of agreement with expert annotations, which is lower than the 0.65 score achieved by Version \ref{enum:version3}. Since Version \ref{enum:version4} differs from Version \ref{enum:version3} only in the model used, the decline is attributable to the switch from ``gpt-4'' to ``gpt-4o''. Figure \ref{fig:chatgpt_version4} illustrates that Version \ref{enum:version4} had 41 instances where experts identified comments as targeting but ChatGPT did not, compared to only 16 in Version \ref{enum:version3}. It correctly identified 95 ``targeting'' comments in line with experts, whereas Version \ref{enum:version3} identified 120. The number of instances where both the model and experts agreed on comments being ``not targeting'' rose from 85 in Version \ref{enum:version3} to 96 in Version \ref{enum:version4}, while the cases where ChatGPT wrongly classified ``not targeting'' comments as ``targeting'' fell from 27 in Version \ref{enum:version3} to 16 in Version \ref{enum:version4}. Version \ref{enum:version4} is thus more conservative than Version \ref{enum:version3} in both directions: it produces fewer false positives but misses considerably more targeting comments, and the net effect on agreement is negative. Model recency and annotation quality are therefore not monotonically related for this task.

\begin{figure}[h!]
    \centering
    \includegraphics[width=\textwidth]{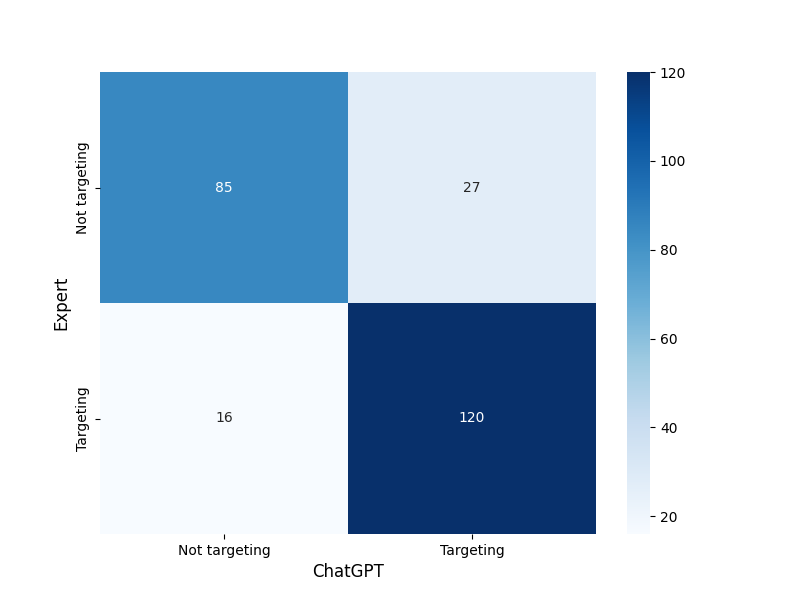}
    \vspace{10pt}
    \caption{Comparison of the Performance of ChatGPT Version 5 in Targeting Language Identification to the Expert Set}
    \label{fig:chatgpt_version5}
\end{figure}

\paragraph{Version \ref{enum:version5}} Version \ref{enum:version5} reverted to the ``gpt-4'' model and lowered the temperature parameter to 0.3. Cohen's Kappa returns to 0.65 from Version \ref{enum:version4}'s 0.54. Relative to Version \ref{enum:version4}, Figure \ref{fig:chatgpt_version5} shows that the number of comments classified as ``TARGETING'' by experts but not by ChatGPT decreased from 41 to 16, and the number of consistently classified ``TARGETING'' comments rose from 95 to 120; the number of consistently classified ``NOT TARGETING'' comments decreased from 96 to 85, and false positives increased from 16 to 27. All four counts are identical to those of Version \ref{enum:version3}, as is the Kappa score. The recovery is therefore attributable to the model reversion rather than to the reduction in temperature, which produced no measurable change in agreement.

\begin{figure}[h!]
    \centering
    \includegraphics[width=\textwidth]{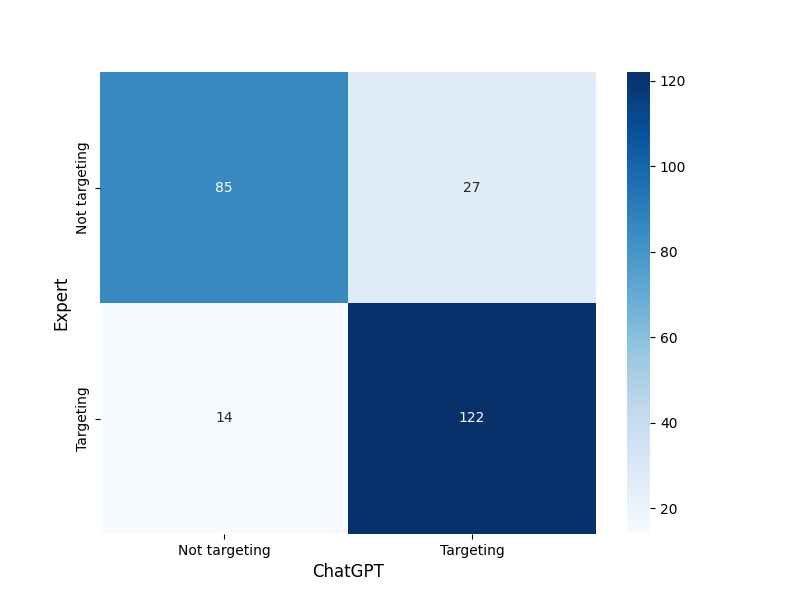}
    \vspace{10pt}
    \caption{Comparison of the Performance of ChatGPT Version 6 in Targeting Language Identification to the Expert Set}
    \label{fig:chatgpt_version6}
\end{figure}

\paragraph{Version \ref{enum:version6}} Comparing Version \ref{enum:version6} to Version \ref{enum:version5}, the Cohen's Kappa score improved slightly from 0.65 to 0.66, indicating a modest increase in agreement with human experts. Figure \ref{fig:chatgpt_version6} shows a reduction in the number of comments labeled as ``TARGETING'' by experts but not by ChatGPT, decreasing from 16 to 14. Additionally, Version \ref{enum:version6} increased the count of correctly identified ``TARGETING'' comments from 120 to 122, while maintaining the number of correctly labeled ``NOT TARGETING'' comments at 85 and the number of false positives at 27. Setting the temperature to 0 therefore produced a small gain, confined to comments the previous configurations had missed.

\par The iterative refinement of prompts and model settings across Versions \ref{enum:version1} through \ref{enum:version6} shows where the gains in this task actually come from. Version \ref{enum:version1}, which simplified the label scheme and moved to a more advanced model, accounts for almost the whole improvement, raising Cohen's Kappa from 0.4 to 0.62. The prompt revision carried through Version \ref{enum:version3} added a further 0.03 by trading recall for precision. The model switch in Version \ref{enum:version4} led to a decline that Version \ref{enum:version5} reversed exactly, and the temperature reductions in Versions \ref{enum:version5} and \ref{enum:version6} moved agreement by at most 0.01. Once the task is specified clearly and the model is fixed, in other words, further tuning contributes little. Overall, these results underscore the importance of careful prompt design and model selection in achieving reliable language identification, with Version \ref{enum:version6} representing the most accurate and consistent performance in aligning with expert assessments.

\subsection{Analysis of All Data}
\begin{figure}[h!]
    \centering
    \includegraphics[width=\textwidth]{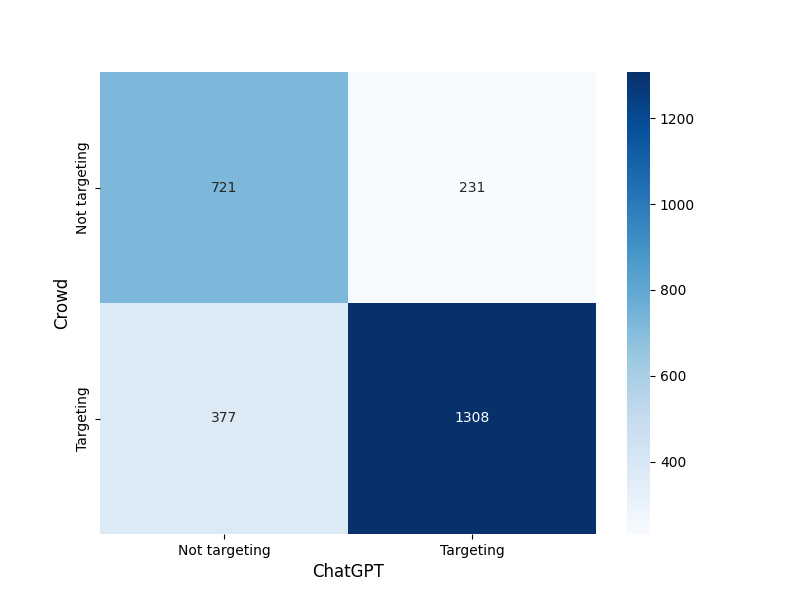}
    \vspace{10pt}
    \caption{Performance Comparison of ChatGPT Version 6 in Targeting Language Identification Against the Crowd Annotations Across All Data}
    \label{fig:chatgpt_version6_crowd}
\end{figure}
\par Given that Version \ref{enum:version6} achieved the highest Cohen's Kappa score of 0.66 in targeting language identification by ChatGPT compared to the expert annotations on the gold set, we applied the same model and prompts to the broader data set. The comparison between ChatGPT's annotations and crowd annotations on all data yielded a Cohen's Kappa score of 0.516, reflecting moderate agreement. This score indicates significant alignment but also reveals discrepancies. Specifically, as seen in Figure \ref{fig:chatgpt_version6_crowd}, ChatGPT missed 377 ``targeting'' comments identified by the crowd and over-identified 231 ``targeting'' comments that the crowd did not recognize. However, both ChatGPT and the crowd agreed on 1308 ``targeting'' comments and 721 ``not targeting'' comments, demonstrating the model's overall reliability. Of the 2,637 units, 39 received no usable label from the model and were counted as targeting when the confusion matrix was constructed, so a portion of the 231 units ChatGPT marked as targeting without crowd agreement reflects unparsed responses rather than substantive disagreement. The residual disagreement runs in both directions rather than reflecting a simple bias toward over-detection. These results highlight Version \ref{enum:version6}'s improvements in annotation accuracy and consistency, while also suggesting areas for further refinement to reduce false negatives and false positives.
\begin{figure}[h!]
    \centering
    \includegraphics[width=\textwidth]{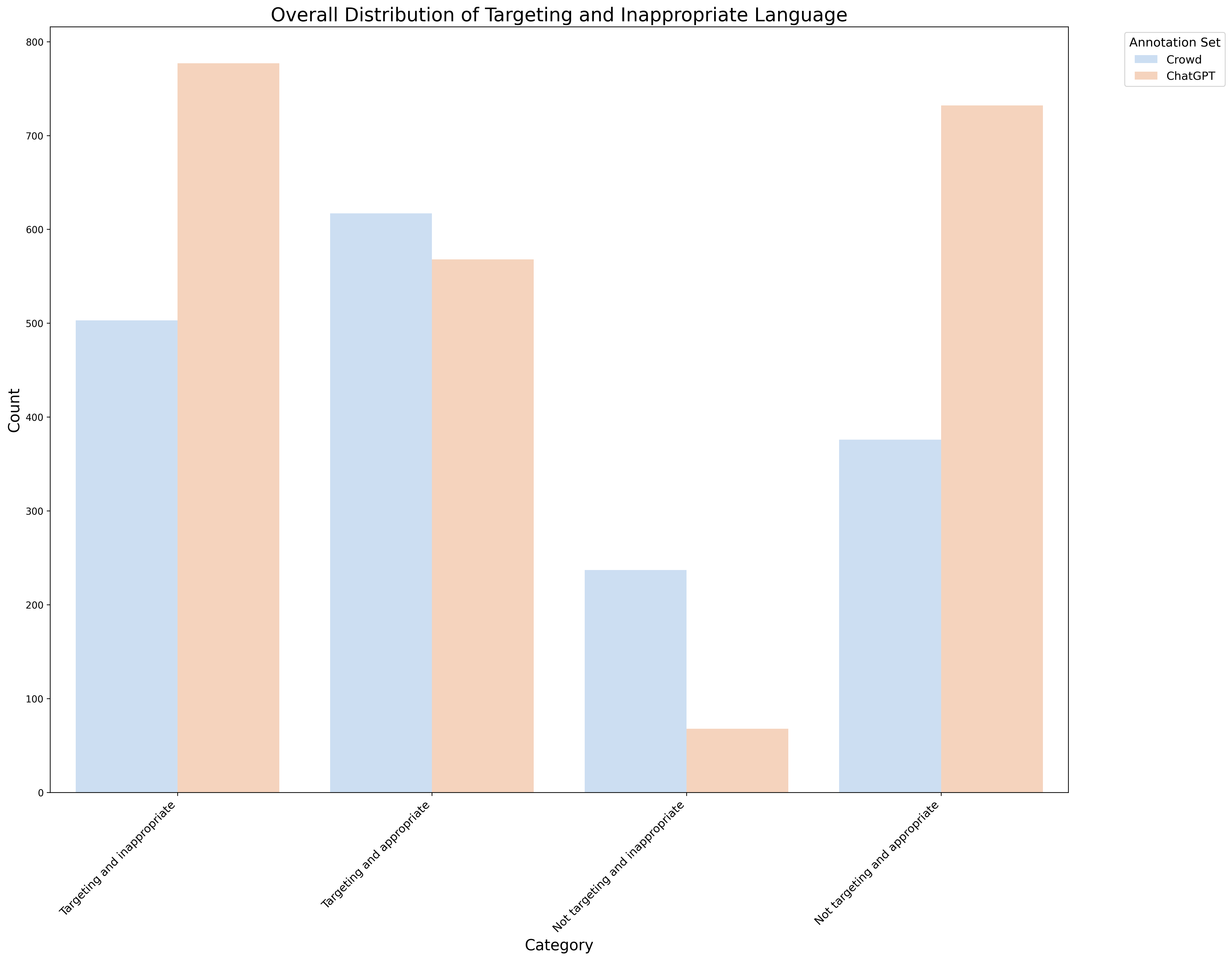}
    \vspace{10pt}
    \caption{ChatGPT vs. Crowd: Overall Distribution of Targeting and Inappropriate Language on All Data}
    \label{fig:overall_distribution}
\end{figure}
\par Building on these findings, a deeper analysis of the distribution of targeting and inappropriate language across ChatGPT and crowd annotations reveals distinct patterns. Figure \ref{fig:overall_distribution} shows that ChatGPT identifies considerably more instances of targeting and inappropriate language (777) than the crowd data set does (503). This over-identification by ChatGPT may contribute to the moderate Cohen's Kappa score observed, indicating that while the model is highly sensitive to potentially targeting language, it may also capture false positives. On the other hand, the crowd data set records higher instances of targeting and appropriate language (617) and non-targeting and inappropriate language (237) than ChatGPT, suggesting that crowd annotations might be more conservative or contextually nuanced.
\begin{figure}[h!]
    \centering
    \includegraphics[width=\textwidth]{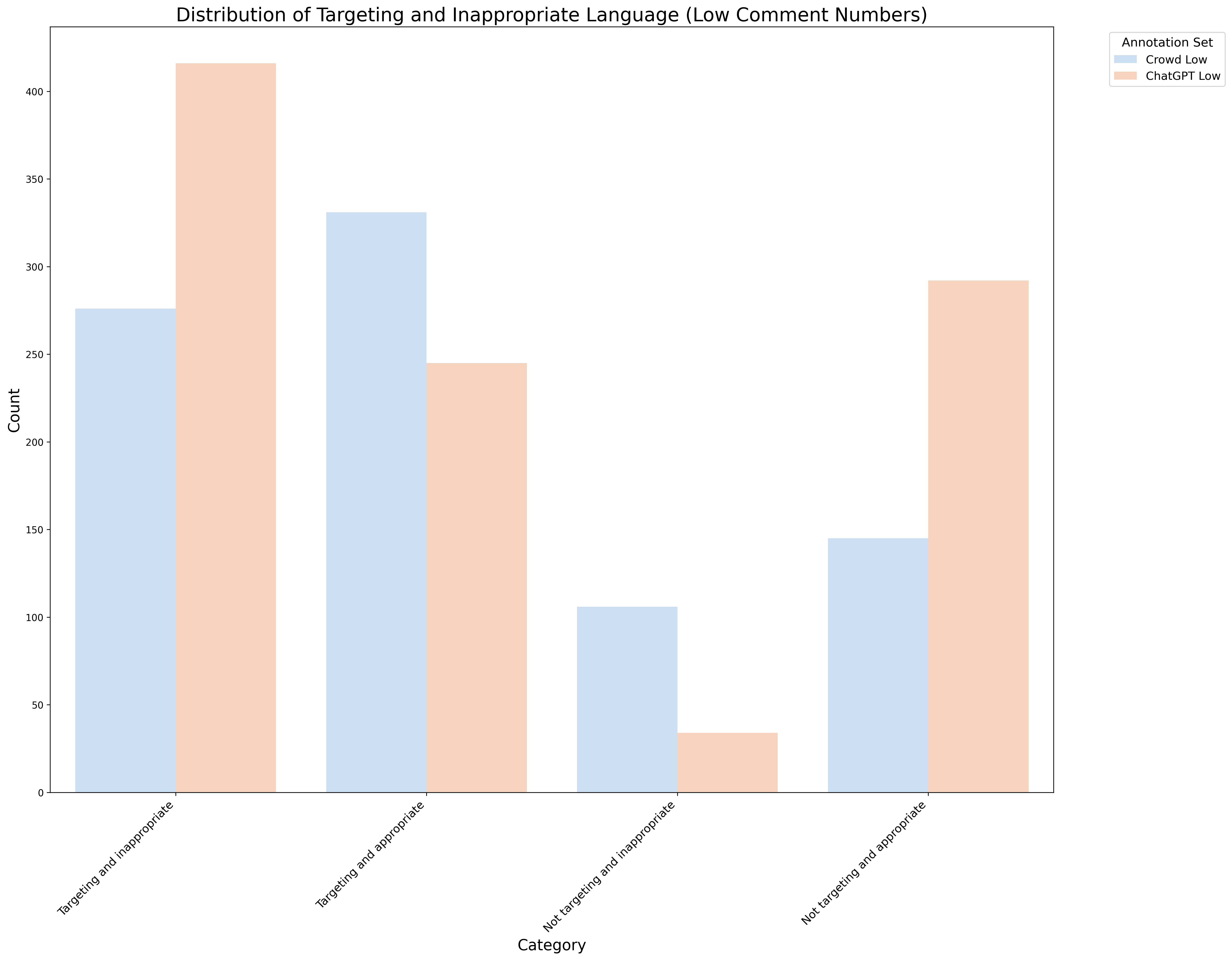}
    \vspace{10pt}
    \caption{ChatGPT vs. Crowd: Distribution of Targeting and Inappropriate Language (Low Comment Numbers) on All Data}
    \label{fig:low_comment_distribution}
\end{figure}
\begin{figure}[h!]
    \centering
    \includegraphics[width=\textwidth]{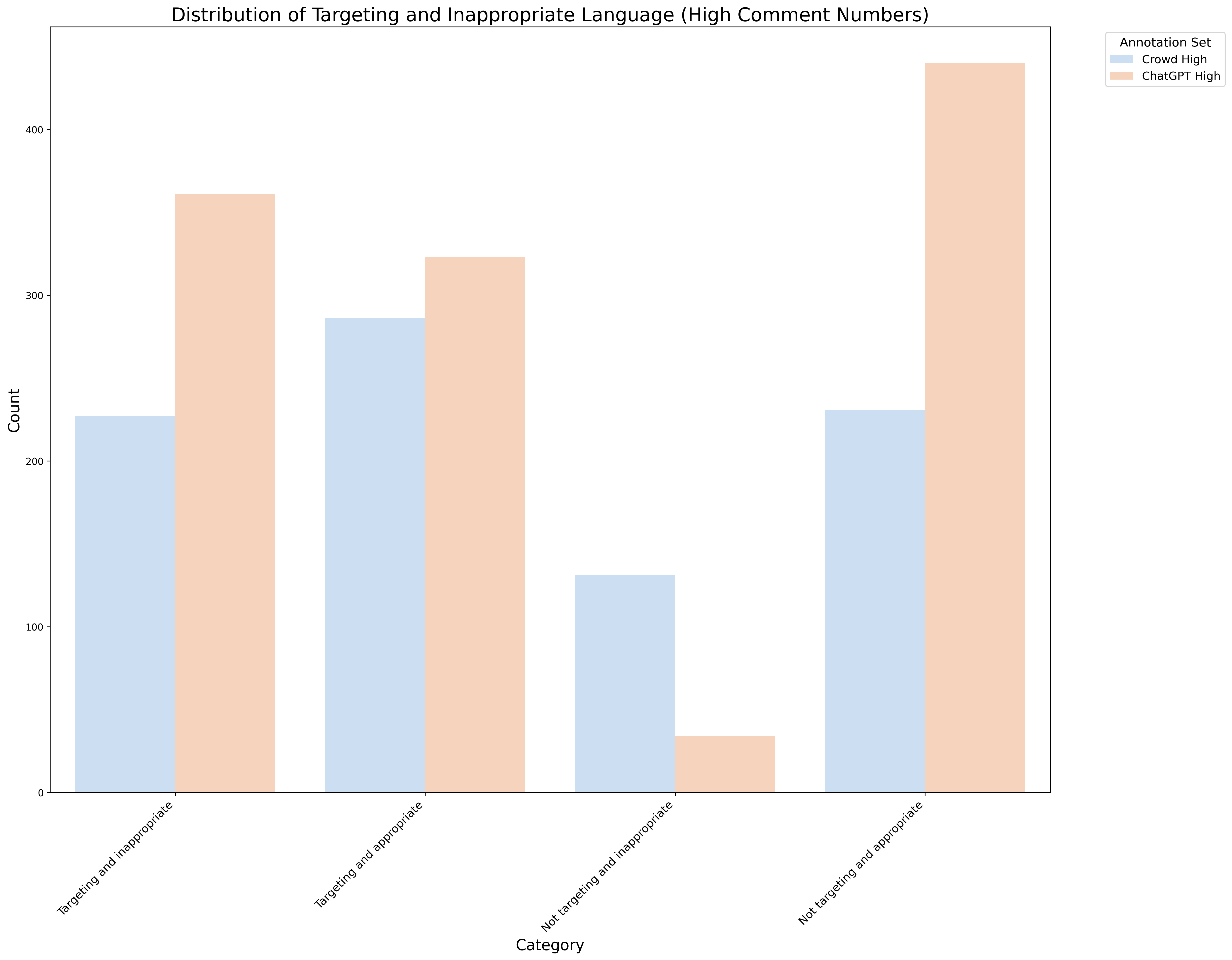}
    \vspace{10pt}
    \caption{ChatGPT vs. Crowd: Distribution of Targeting and Inappropriate Language (High Comment Numbers) on All Data}
    \label{fig:high_comment_distribution}
\end{figure}
\par Figures \ref{fig:low_comment_distribution} and \ref{fig:high_comment_distribution} both highlight interesting trends. Since the minimum number of comments per conversation thread is 3, we used a threshold to differentiate between low and high comment numbers, with ``low'' referring to the first two comments (less than three) and ``high'' referring to comments from the third onwards (three or more). In the crowd data set, targeting and inappropriate comments are relatively stable between low and high comment numbers, but there is a noticeable shift toward non-targeting and appropriate language in the later comments. This shift implies that conversations tend to become more civil over time. ChatGPT's annotations show a marked decline in targeting and inappropriate language from low to high comment numbers, indicating that while ChatGPT might identify conversations as starting more aggressively, it also recognizes a trend toward more appropriate interactions as discussions progress. The discrepancies in the patterns between the gold set and the broader data set arise from differences in sample size and variability as well as the changes in prompt design. The gold set, with its smaller, curated sample, may reflect more controlled patterns of targeting and inappropriateness, while the broader data set, with its larger and more diverse collection, can exhibit different trends due to increased variability. While the updated prompts improved detection, they also introduced variability in results. This variation highlights the impact of prompt design on annotation outcomes and underscores the need for consistent and precise criteria in evaluating targeting and inappropriateness across different data sets.
\par The analysis of ChatGPT's performance, particularly with Version \ref{enum:version6}, suggests that the model is capable of supporting automated content moderation systems, given its accuracy and consistency in identifying inappropriate language. Its agreement on targeting remains moderate, so it is better suited to complementing human annotators than to replacing them. With improvements in detecting and addressing harmful behavior in real time, such a system could enable early detection of problematic interactions through alert mechanisms and preventive interventions. However, to ensure effectiveness, ongoing refinement of the model, alongside addressing ethical considerations and practical challenges, will be essential.
\section{Conclusion}
\label{sec:conclusion}
\par This study assessed ChatGPT's effectiveness in identifying targeting and inappropriate language in online comments, revealing both its strengths and its limits. ChatGPT proved capable in detecting inappropriate content, reaching a Cohen's Kappa of 0.88 against adjudicated expert annotations at the comment level and agreeing with expert judgments on 16 of 17 and 28 of 29 units in the two scenarios where it was asked about inappropriateness without any labels supplied. Its detection of targeting language was weaker and less stable throughout, with high false positive rates before prompt refinement and high false negative rates after a change of model family.
\par Three findings bear on how such systems should be built and evaluated. First, the gap between the two tasks is not a matter of how much context the model is given. Where the model received identical input and was asked both questions of the same comment, it agreed with the experts on inappropriateness in 16 of 17 cases but on targeting in only 8 of 17. The difficulty therefore lies in the relational nature of the targeting judgment --- deciding whom a comment positions as its object --- rather than in the information available at the point of decision. That the human annotators show the same asymmetry, with expert agreement on targeting reaching only 0.58 against 0.76 on inappropriateness, suggests the difficulty is intrinsic to the judgment rather than particular to the model.
\par Second, almost the entire gain from iterative refinement came from a single change. Simplifying the label scheme from a three-way distinction to a binary one and moving from \texttt{text-davinci-003} to \texttt{gpt-4} raised Cohen's Kappa from 0.4 to 0.62, and the counts show that the improvement came entirely from better recognition of comments that were \textit{not} targeting: shared \textit{not targeting} judgments rose from 51 to 75 and false positives fell from 61 to 37, while shared \textit{targeting} judgments were unchanged at 127. Subsequent revisions to prompt structure, logging and error handling, and the reduction of sampling temperature from 0.8 to 0.3 and then to 0, together moved agreement from 0.62 to 0.66. The practical implication is that effort is better spent on specifying the task clearly than on incremental prompt engineering.
\par Third, model recency and annotation quality are not monotonically related. Substituting \texttt{gpt-4o} for \texttt{gpt-4} while holding the prompt fixed reduced agreement from 0.65 to 0.54, with missed targeting comments rising from 16 to 41; reverting the model restored the earlier figures exactly. Anyone deploying such a system should therefore treat a model upgrade as a change requiring revalidation rather than as an improvement by default.
\par Applied to the broader data set, the best configuration reached a Cohen's Kappa of 0.516 against crowd annotations, with disagreement running in both directions rather than reflecting a simple bias toward over-detection. These findings suggest that ChatGPT has considerable potential to support or complement human moderators in automated content moderation systems, offering a more scalable approach to managing online interactions, but they do not support replacing human judgment on targeting, where agreement with expert annotation remains moderate at best. Further refinement is needed to improve targeting detection and to reduce both false positives and false negatives. Future research should address these aspects, establish how much of the residual disagreement reflects genuine interpretive difficulty rather than model error, and continue to attend to the ethical considerations that arise when moderation decisions are made automatically.
\section{Limitations}
\label{sec:limit}
\par This study has several limitations that must be acknowledged, some concerning the model evaluated and some concerning the evaluation itself.
\par The first set concerns scale. The contextual-cue evaluation reported in Table \ref{tab:chatgpt_results} rests on small subsets: 14 subthreads for Scenarios \ref{enum:scenario1} and \ref{enum:scenario2}, 17 for Scenario \ref{enum:scenario3}, and 29 for Scenario \ref{enum:no labels context last}. At these sizes a single comment moves the reported percentage by between three and seven points, so only substantial differences can be interpreted, and the comparison between Scenarios \ref{enum:scenario1} and \ref{enum:scenario2} in particular turns on one comment. The prompt-refinement analysis rests on a larger but still limited base of 248 annotated units drawn from 39 subthreads. More generally, the gold set was drawn by stratified sampling on lexicon-based toxicity score and number of comments from subthreads that were themselves selected on lexicon matches, so the data used for evaluation over-represent inappropriate language and may not fully capture the diversity and complexity of real-world online interactions, which limits the generalizability of the findings.
\par A second set concerns the design of the prompts and the recording of the model's answers. In Scenario \ref{enum:scenario1} the context is supplied with a label convention under which \texttt{'targeting\_label': False} marks a comment as targeting and \texttt{True} marks it as not targeting, while the answer format for the question actually asked uses \texttt{True} to mean that the comment \textit{is} inappropriate. Scenarios \ref{enum:scenario3} and \ref{enum:no labels context last} ask both questions in one prompt with opposite conventions: \texttt{True\_I} marks a comment as inappropriate, whereas \texttt{False\_T} marks it as targeting. Scenario \ref{enum:scenario1} therefore cannot cleanly separate the contribution of contextual information from the effect of label polarity, and in Scenarios \ref{enum:scenario3} and \ref{enum:no labels context last} part of the gap between agreement on inappropriateness and on targeting may reflect the inverted label rather than the difference between the two judgments. The model's replies were converted into labels by searching them for the answer strings, rather than by constraining the output format, and the reply length was capped at 50 tokens. In Scenarios \ref{enum:scenario1} and \ref{enum:scenario2}, any reply that does not contain the word \textit{false} is counted as inappropriate; in Scenarios \ref{enum:scenario3} and \ref{enum:no labels context last}, a reply that contains neither answer string keeps the value recorded for the previous comment. A hedged, refused, or truncated answer is therefore not flagged as unusable. In the full-data Version 6 run, 39 of 2,637 units returned no usable label and were counted as targeting, so a small part of the reported false positive rate is attributable to unparsed responses rather than to misclassification.
\par A third set concerns reproducibility. The evaluations were run against the \texttt{gpt-4} and \texttt{gpt-4o} endpoints without pinning a dated model snapshot or fixing a random seed, and Versions \ref{enum:version1} through \ref{enum:version4} were run at a sampling temperature of 0.8. Each configuration was evaluated once, so we cannot estimate run-to-run variance, and small differences between adjacent versions should not be read as effects of the changes those versions introduced. Because the underlying endpoints are revised over time, exact replication of these figures is unlikely even with identical prompts.
\par A fourth set concerns the reference standards. The expert annotations that serve as the benchmark for the gold set were produced by the authors of this study, which introduces the possibility that the annotation guidelines and the expert judgments share assumptions that an independent annotator would not hold. The crowd annotations used for the large-scale comparison were produced by non-expert annotators and show only moderate agreement with the experts on targeting, so discrepancies between ChatGPT and the crowd cannot be attributed to the model alone. Agreement figures throughout should be read as measures of alignment between annotation sources rather than as measures of accuracy against a ground truth.
\par Finally, several limitations concern the model's behavior and its deployment. ChatGPT's performance in identifying targeting language exhibited variability, with higher false positive rates than expert judgments in the earlier prompt versions and higher false negative rates after the switch to \texttt{gpt-4o}. This suggests that while ChatGPT is effective in detecting inappropriate content, its ability to classify nuanced targeting language requires further refinement, and that its performance remains dependent on prompt design and model parameters in ways that may not transfer to other contexts or languages. Ethical considerations and the potential for biased behavior in AI models likewise remain concerns that need to be addressed to ensure the fair and responsible use of automated content moderation tools, particularly where a false positive results in a moderation action against a user.
\section{Acknowledgments}
This research was supported by Huawei Finland through the DreamsLab project. All content represents the opinions of the authors, which are not necessarily shared or endorsed by their respective employers and/or sponsors.
\appendix
\section{ChatGPT Prompts for Targeting and Inappropriate Language Detection With/Without Contextual Cues}
\label{appendix:chatgpt_scenarios}
\par This appendix provides detailed descriptions of the scenarios used in the evaluation of ChatGPT's performance in identifying inappropriate and targeting comments using experts' original labels. Figures \ref{fig:scenario1}, \ref{fig:scenario2}, and \ref{fig:scenario3,4} present the specific scenarios with the corresponding prompts used to guide ChatGPT's responses. All four scenarios were run against the \texttt{gpt-4} endpoint. Scenarios 1 and 2 ask the same question of the same comment and differ only in whether the targeting labels of the surrounding comments are supplied, so their prompts are identical apart from the explanatory system message and the presence of the \texttt{targeting\_label} field in the serialized context. Scenarios 3 and 4 share a single prompt, shown in Figure \ref{fig:scenario3,4}, and differ only in which comment of the thread is selected for classification.
\begin{landscape}
\begin{figure}[h]
    \centering
    \resizebox{\textwidth}{!}{
    \fbox{\parbox{\textwidth}{
    \scriptsize

    messages=[
                {"role": "system", "content": "Note: "'targeting\_label': False" indicates that the comment is targeting a group or individual while "'targeting\_label': True" indicates that the comment is not targeting a group or individual. Here are the previous comments in the conversation and their relevant data:"},

                {"role": "user", "content": previous\_comments},

                {"role": "system", "content": "Last comment and data:"},

                {"role": "user", "content": last\_comment\_str},

                {"role": "user", "content": f"Is comment {comment} explicitly inappropriate? Say True if it is explicitly inappropriate. Say False if it is not explicitly inappropriate."}
            ]
    }}
    }
    \caption{Scenario 1}
    \label{fig:scenario1}
\end{figure}

\begin{figure}[h]
    \centering
    \resizebox{\textwidth}{!}{
    \fbox{\parbox{\textwidth}{
    \scriptsize

    messages=[
                {"role": "system", "content": "Note: Here are the previous comments in the conversation and their relevant data:"},

                {"role": "user", "content": previous\_comments},

                {"role": "system", "content": "Last comment and data:"},

                {"role": "user", "content": last\_comment\_str},

                {"role": "user", "content": f"Is comment {comment} explicitly inappropriate? Say True if it is explicitly inappropriate. Say False if it is not explicitly inappropriate."}
            ]
    }}
    }
    \caption{Scenario 2. The \texttt{targeting\_label} column is removed from the serialized context before the prompt is built, which is what distinguishes this scenario from Scenario 1.}
    \label{fig:scenario2}
\end{figure}

\begin{figure}[h]
    \centering
    \resizebox{\textwidth}{!}{
    \fbox{\parbox{\textwidth}{
    \scriptsize

    messages=[
                {"role": "system", "content": "Note: Here are the previous comments in the conversation and their relevant data:"},
                {"role": "user", "content": previous\_comments},

                {"role": "system", "content": "Last comment and data:"},

                {"role": "user", "content": last\_comment\_str},

                {"role": "user", "content": f"Is comment {comment} explicitly inappropriate? Say True\_I if it is explicitly inappropriate. Say False\_I if it is not explicitly inappropriate. Is comment {comment} targeting a group or individual? Say False\_T if it is targeting. Say True\_T if it is not targeting."}
            ]
    }}
    }
    \caption{Scenarios 3 and 4}
    \label{fig:scenario3,4}
\end{figure}
\end{landscape}
\section{Evolution of ChatGPT Prompts for Targeting Language Detection}
\label{appendix:chatgpt_improvements}
\par This appendix presents the various versions of prompts used to guide ChatGPT in annotating Reddit comments for targeting behavior. Each version of the prompt was designed to refine and improve the accuracy of the model's classification. The evolution of these prompts is documented to illustrate the iterative process of enhancing the model's performance. While Figure \ref{fig:original_prompt} demonstrates the original prompt used by \citet{barbarestani2024content}, Figures \ref{fig:version1}, \ref{fig:version2}, and \ref{fig:version3} show the different prompt versions, detailing the instructions given to the model and the examples provided to facilitate accurate annotation. Versions 3 through 6 share the prompt shown in Figure \ref{fig:version3} and differ only in the model and the sampling temperature. Offensive words are obfuscated here, though they were not in the actual prompts.
\begin{enumerate}
    \item \textbf{Version 1:} This version introduces several changes aimed at refining the identification of targeting language in comments. Unlike the original version, which used a prompt focused on categorizing comments into specific targeting types (``I'' for targeting inside the conversation thread, ``O'' for targeting outside the conversation thread, and ``N'' for not targeting), Version 1 simplifies the categorization into a binary distinction of ``TARGETING'' or ``NOT TARGETING''. This refinement narrows the model's focus by reducing the complexity of the decision-making process. Additionally, Version 1 transitions from using the ``text-davinci-003'' model to ``gpt-4''. The new code optimizes the handling of responses for clarity and consistency by using the ``gpt-4'' chat-based model, ensuring that the output is clear and less ambiguous. The implementation in Version 1 further streamlines the process by using the ChatCompletion endpoint.
    \item \textbf{Version 2:} This version refines the approach further by introducing a modular structure and enhanced error handling. Unlike Version \ref{enum:version1}, which directly constructs and manages prompts within the main processing loop, Version 2 separates prompt generation into its own function, improving code organization and readability. It also incorporates a retry mechanism with a maximum retry limit to handle service disruptions more gracefully, a feature not present in Version \ref{enum:version1}. Additionally, this version provides more detailed error handling by logging failures when responses cannot be obtained after multiple retries. This version also rewrites the instructions as numbered steps and asks the model to respond only with ``TARGETING'' or ``NOT TARGETING'' (Figure \ref{fig:version2}).
    \item \textbf{Version 3:} This version distinguishes itself from Version \ref{enum:version2} by incorporating enhanced logging and error management. It introduces logging to track the processing stages and capture errors for transparency and debugging. This version also refines error handling in API interactions, with specific handling for various types of errors and a more robust retry mechanism. Additionally, this version refactors the code into modular functions, for readability and maintainability, by clearly separating different tasks such as generating prompts, classifying responses, and processing CSV files. This version also rewrites the instructions, asking whether the text clearly targets or attacks a specific person or group (Figure \ref{fig:version3}); Versions \ref{enum:version4} to \ref{enum:version6} keep this prompt.
    \item \textbf{Version 4:} This version is similar to Version \ref{enum:version3} in structure and functionality but uses the ``gpt-4o'' model instead of ``gpt-4''. This adjustment reflects a shift to a different version of the model. The code maintains robust logging practices and error handling mechanisms, ensuring comprehensive tracking of processing steps and issues. It also continues to read and process CSV files similarly to Version \ref{enum:version3}, handling each file by generating prompts and analyzing responses based on the predefined criteria. The overall approach to classifying text as ``TARGETING'' or ``NOT TARGETING'' remains consistent, with the primary difference being the updated model version.
    \item \textbf{Version 5:} This version introduces a key adjustment by reverting to the ``gpt-4'' model instead of ``gpt-4o'', while also lowering the temperature parameter from 0.8 to 0.3, for more deterministic and consistent responses. This version maintains the same overall structure and approach for processing and classifying Reddit comments as either ``TARGETING'' or ``NOT TARGETING''. The handling of CSV files, logging, and error management remains consistent with Version \ref{enum:version4}.
    \item \textbf{Version 6:} This version primarily differs from Version \ref{enum:version5} in its adjustment of the GPT model's temperature setting, which has been changed from 0.3 to 0, for more consistent and less varied responses from the model by making the output more deterministic. The rest of the code remains the same, including the setup for logging, error handling, and the process for reading from and writing to CSV files. The script still uses the ``gpt-4'' model and performs the same tasks, but with an emphasis on obtaining more predictable and uniform responses.
\end{enumerate}
\begin{landscape}
\begin{figure}[h]
    \centering
    \resizebox{\textwidth}{!}{
    \fbox{\parbox{\textwidth}{
    \scriptsize

    You are a language model that helps annotate inappropriate words and harmful messages in comments.
    Given a CSV file, First, you should read the following columns per row: "title\_text", "comment", "context", "user\_id", and "post\_text".
    "context" refers to the previous comments.
    You should read each "title\_text", "post\_text", and "context". Then, you should read "comment".
    If "comment" is not "-", you should produce your response based on "comment". If "comment" is "-", you should produce your response based on "title\_text".

    Instructions

    You will review comments/titles posted on Reddit that may contain inappropriate language. Your task is to identify if they target a specific person or group of people. You should first read the post text. Read the comment/title and assess whether it targets anyone.
    If the comment/title does not target someone, mark the comment/title as "not targeting". If the comment/title targets someone, you should decide whether the comment/title is targeting a person inside the conversation thread or outside of the conversation thread.
    Mark targeting a person inside the conversation thread as I.
    Mark targeting a person outside of the conversation thread as O.
    Mark not targeting as N.

    Examples:

    I: Sit the f**k down, you colossal piece of sh*t.

    I: Stupid me did not press the record button.

    O: They should all be ashamed of themselves.

    N: The sky is clear and the sun is shining brightly today.

    N: This f**king life is crazy!
    }}
    }
    \caption{Original Prompt}
    \label{fig:original_prompt}
\end{figure}

\begin{figure}[h]
    \centering
    \resizebox{\textwidth}{!}{
    \fbox{\parbox{\textwidth}{
    \scriptsize

    You are a language model that helps annotate inappropriate words and harmful messages in comments.
    Given a CSV file, you should read the following columns per row: "title\_text", "comment", "context", "user\_id", and "post\_text".
    "context" refers to the previous comments.

    You should read each "title\_text", "post\_text", and "context". Then, you should read "comment".

    If "comment" is not "-", you should produce your response based on "comment". If "comment" is "-", you should produce your response based on "title\_text".

    Instructions:
    You will review comments/titles posted on Reddit that may contain inappropriate language. Your task is to identify if they target a specific person or group of people inappropriately. You should first read the post text. Read the comment/title and assess whether it targets anyone.

    If the comment/title targets someone, mark it as "TARGETING". If the comment/title does not target someone, mark it as "NOT TARGETING".

    Examples:

    TARGETING: Sit the f**k down, you colossal piece of sh*t.

    TARGETING: They should all be ashamed of themselves.

    NOT TARGETING: The sky is clear and the sun is shining brightly today.

    NOT TARGETING: This f**king life is crazy!
    }}
    }
    \caption{Prompt Version 1}
    \label{fig:version1}
\end{figure}
\end{landscape}

\begin{landscape}
\begin{figure}[h]
    \centering
    \resizebox{\textwidth}{!}{
    \fbox{\parbox{\textwidth}{
    \scriptsize

    You are an AI language model that helps annotate Reddit comments for inappropriate or harmful content. You will receive a comment or title, along with its context, and your task is to determine whether it targets a specific person or group.

    Instructions:

    1. First, read the "post\_text" and "context" to understand the background.

    2. Then, analyze the "comment" or "title\_text":

       - If "comment" is "-", assess the "title\_text".

       - Otherwise, assess the "comment".

    3. Determine if the comment/title specifically targets or attacks an individual or group.

    Respond only with one of the following labels:

    - "TARGETING" if it targets someone.

    - "NOT TARGETING" if it does not target anyone.

    Examples:

    - TARGETING: "Sit the f**k down, you colossal piece of sh*t."

    - TARGETING: "They should all be ashamed of themselves."

    - NOT TARGETING: "The sky is clear and the sun is shining brightly today."

    - NOT TARGETING: "This f**ing life is crazy!"

    Remember, your response must be either "TARGETING" or "NOT TARGETING".
    }}
    }
    \caption{Prompt Version 2}
    \label{fig:version2}
\end{figure}

\begin{figure}[h]
    \centering
    \resizebox{\textwidth}{!}{
    \fbox{\parbox{\textwidth}{
    \scriptsize

    You are a highly capable language model tasked with identifying whether comments or titles on Reddit target a specific person or group of people inappropriately. Your job is to classify each comment or title as either "TARGETING" or "NOT TARGETING" based on the provided context.

    \#\#\# Instructions:

    1. **Read the Content**: You will review the "title\_text", "post\_text", and "context" provided in the input.

    2. **Identify the Relevant Text**: Focus on the "comment" if it is not "-", otherwise, focus on the "title\_text".

    3. **Classify the Comment/Title**:

       - If the text in the comment/title clearly targets or attacks a specific person or group, classify it as "TARGETING".

       - If the text does not target anyone specifically, classify it as "NOT TARGETING".

    \#\#\# Examples:

    - **TARGETING**: "Sit the f**k down, you colossal piece of sh*t."

    - **TARGETING**: "They should all be ashamed of themselves."

    - **NOT TARGETING**: "The sky is clear and the sun is shining brightly today."

    - **NOT TARGETING**: "This f**king life is crazy!"

    \#\#\# Your Task:

    Based on the above instructions, classify the following comment/title as either "TARGETING" or "NOT TARGETING".
    }}
    }
    \caption{Prompt Versions 3, 4, 5, and 6}
    \label{fig:version3}
\end{figure}
\end{landscape}
\section{Ethics Statement}
The research presented in this paper is conducted with a strong commitment to ethical standards and responsible AI use. The data used were originally anonymized, ensuring that individual identities are protected. The study aims to enhance automated content moderation systems while acknowledging the potential biases inherent in both human and AI annotations. We are committed to transparency about these biases and to minimizing their impact through iterative model improvements and rigorous validation processes. Ethical considerations are central to our work, and we remain dedicated to using our findings to support fair and equitable online interactions, while continually addressing and mitigating any unintended consequences of our technology.
 \bibliographystyle{nlplike}
 \bibliography{Sample}

\label{lastpage}

\end{document}